%% file: arxiv.tex
\documentclass[10pt,twocolumn,letterpaper]{article}
\usepackage{cvpr}
\input{preamble}
\definecolor{cvprblue}{rgb}{0.21,0.49,0.74}
\usepackage[pagebackref,breaklinks,colorlinks,allcolors=cvprblue,hypertexnames=false]{hyperref}
\def\confName{CVPR}
\def\confYear{2027}
\title{SAGE-VQA: A Training-Free, Self-Evolving Agentic Framework with Grounded Evidence for Knowledge-Based Visual Question Answering}
\input{authors}
\begin{document}
\maketitle

\begin{abstract}
Knowledge-based visual question answering (KB-VQA) aims to combine visual information with external knowledge to answer user questions.
Existing methods face two main bottlenecks: visual perception errors can propagate into hallucinations, and generalization across scenarios remains limited.
To address these issues, we propose \method (Self-evolving Agent framework with Grounded Evidence), a training-free agentic framework that combines evidence verification with experience evolution.
Our key insight is to explore an Evidence-Driven Adaptive Reasoning (EDAR) strategy. It accurately acquires and verifies question-relevant fine-grained visual information. It also extracts and refines reusable experience to support robust reasoning across tasks.
Specifically, we design a Co-verified Visual Grounding module that verifies question-relevant visual information and provides grounded evidence for subsequent reasoning.
We further introduce a Self-evolving Hierarchical Memory (SHM) mechanism. SHM organizes experiences into declarative and procedural memories based on their nature.
These memories can be reused in subsequent reasoning and continually refined through feedback, without updating model parameters.
Experimental results show that our method outperforms competing methods based on lightweight open-source vision-language models (VLMs) as well as those based on closed-source VLMs.
Moreover, our method can serve as a plug-and-play reasoning framework for existing VLMs without modifying their architectures.
\end{abstract}

\section{Introduction}
\label{sec:intro}

Knowledge-based visual question answering (KB-VQA) aims to combine images and external knowledge bases to answer user questions~\citep{marino2019okvqa,schwenk2022aokvqa,lu2022scienceqa}.
Images provide visual evidence relevant to the question, and external knowledge bases provide additional information.
Closed-source vision-language models (VLMs) achieve strong performance on VQA tasks due to their powerful reasoning capabilities~\citep{yue2024mmmu}. However, their reliance on remote APIs increases usage costs and limits personalized adaptation and local deployment.
Therefore, improving the knowledge question-answering capabilities of lightweight open-source VLMs that support local deployment has practical value.

Existing KB-VQA methods fall into two broad categories: training-based and training-free methods.
Training-based methods adapt to specific KB-VQA tasks by fine-tuning models or training auxiliary modules~\citep{zhang2024mmcot,yang2025qacap}.
For example, MR-MKG trains a knowledge encoder and cross-modal alignment modules to improve multimodal knowledge reasoning~\citep{lee2024mrmkg}.
These methods are effective on specific datasets but require extensive manual annotation and substantial training costs.
Task-specific fine-tuning also limits their generalization to other tasks and domains.
This motivates improving reasoning in lightweight open-source VLMs without task-specific fine-tuning.

\begin{figure*}[t]
\centering
\includegraphics[
  width=0.8\linewidth,
  trim=0 218bp 0 105bp,
  clip
]{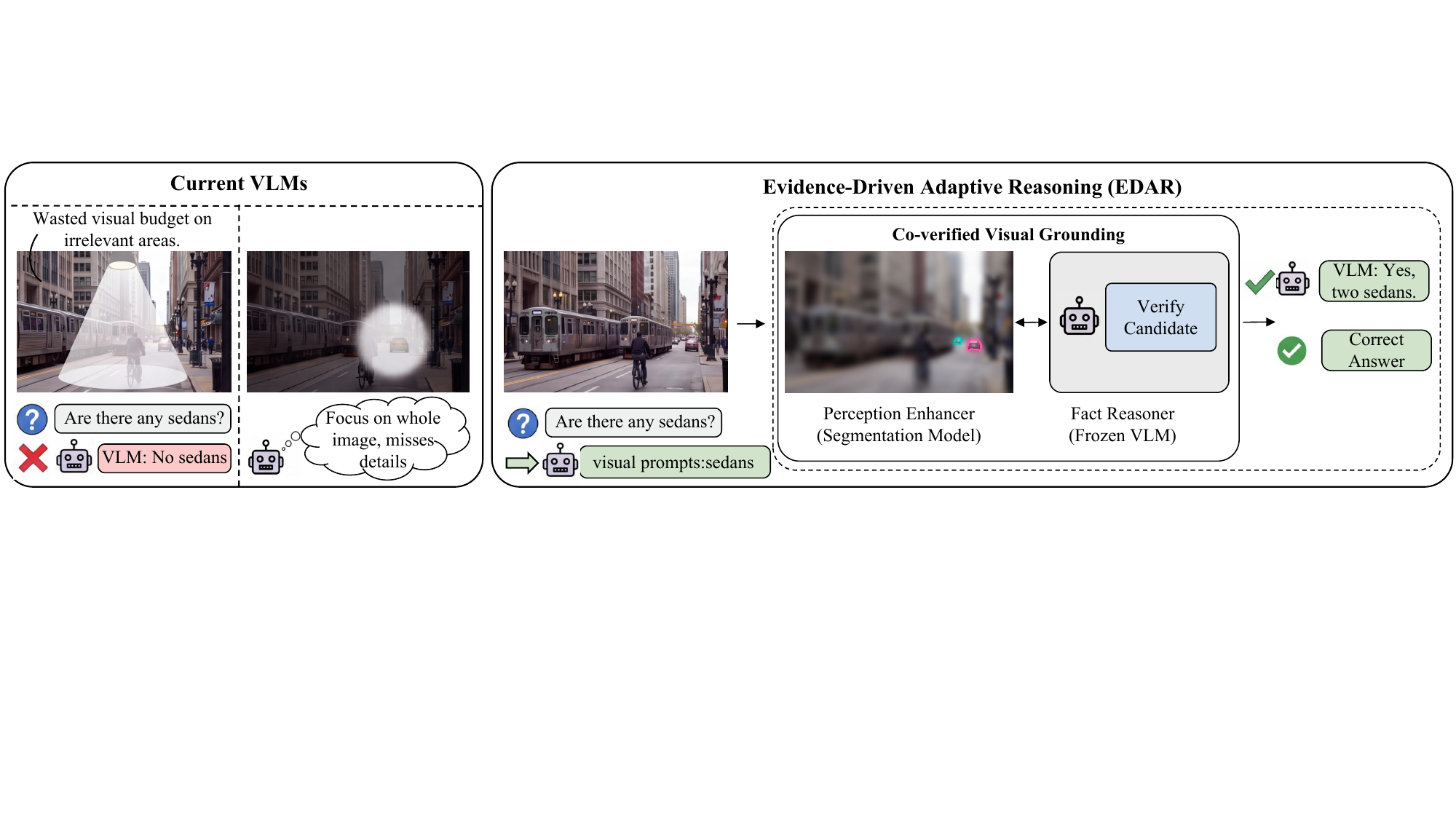}
\caption{Motivation for EDAR. Whole-image processing can allocate visual capacity to question-irrelevant regions and miss answer-relevant local details. Co-verified Visual Grounding locates and verifies candidate regions before answer generation.}
\label{fig:motivation}
\end{figure*}

Training-free methods improve VQA through prompt enhancement and tool use.
For example, PICa combines image captions, tags, and example question--answer pairs into prompts.
It uses the knowledge of VLMs to answer questions~\citep{yang2022pica}.
AVIS dynamically calls tools during reasoning to obtain the information needed for answering~\citep{hu2023avis}.
However, reasoning in these methods still depends on the accuracy of visual information.
Existing VLMs can still produce object hallucinations and fine-grained visual perception errors~\citep{hoang2026pas,li2026groundingme}.
As shown in Figure~\ref{fig:motivation}, a VLM processes the entire image.
Part of its visual processing budget is spent on question-irrelevant objects and backgrounds~\citep{han2023survey,brauwers2023general}.
This limits its attention to key objects.
Whole-image representations may also fail to focus sufficiently on local details relevant to the answer.
These two limitations can cause visual information to be missed or misinterpreted and affect subsequent reasoning.
This raises the first challenge: how can we acquire question-relevant visual evidence without updating model parameters?
The goal is to reduce the impact of perception errors on subsequent reasoning.

In visual question answering, questions of the same type may share similar reasoning experience.
It is therefore important to retain useful knowledge and solution methods for later reasoning on similar questions.
Existing studies store reflections, past experience, or executable skills to support experience reuse in model reasoning~\citep{shinn2023reflexion,park2023generative,wang2023voyager,xu2025amem}.
For example, ReasoningBank distills reusable reasoning strategies from successful and failed task trajectories.
It retrieves relevant experience for later tasks and appends newly extracted memories to its memory bank~\citep{ouyang2026reasoningbank}.
However, past experience may not apply to new, similar questions.
Simply adding more memories therefore creates redundancy and makes accurate retrieval difficult.
It also reduces the applicability of existing experience.
This raises the second challenge: how can a model reuse relevant knowledge and methods for new questions and continuously refine past experience through feedback?
How can this process support accurate memory retrieval and robust use across different question types?

To address these two challenges, we propose \method (Self-evolving Agent framework with Grounded Evidence).
This training-free agentic framework combines evidence verification with experience evolution.
Our key insight is to explore an Evidence-Driven Adaptive Reasoning (EDAR) strategy.
The strategy aims to accurately acquire question-relevant fine-grained visual information and use refined experience to support robust reasoning across tasks.
To address the first challenge, we design a Co-verified Visual Grounding module to verify visual information.
A pretrained segmentation model serves as the perception enhancer and provides question-relevant visual candidates from the image.
A frozen VLM serves as the fact reasoner and checks semantic consistency between the target objects and these visual candidates.
To address the second challenge, we introduce a Self-evolving Hierarchical Memory (SHM) mechanism.
SHM organizes experience by its nature and use into declarative and procedural memory.
Declarative memory includes episodic records of reasoning and corrections, as well as semantic rules drawn from similar episodes.
Procedural memory stores executable methods and their conditions of use.
To initialize the bank, the system extracts reusable experience from selected training cases.
For later questions, the system retrieves relevant memories, and SHM refines the bank using post-answer feedback.
Finally, EDAR combines visual evidence, knowledge from a multimodal knowledge graph, and relevant memories for reasoning.
It assesses readiness at each iteration to determine whether the available evidence is sufficient to answer the user's question.

Our contributions are as follows:
\begin{itemize}
  \item We propose SAGE-VQA, a training-free framework that uses EDAR to combine question-relevant visual verification with self-evolving experience. It improves the KB-VQA\pagebreak[0] performance of lightweight open-source VLMs without task-specific parameter training.
  \item We design Co-verified Visual Grounding, which uses a pretrained segmentation model as a perception enhancer to obtain question-relevant visual candidates and a frozen VLM to assess their semantic consistency with target objects. This strengthens perception of relevant visual details before downstream reasoning.
  \item We introduce SHM to support the reuse and feedback-based refinement of declarative and procedural memories, allowing the framework to improve past experience without updating model parameters.
  \item We evaluated SAGE-VQA as a plug-and-play reasoning framework on four datasets. The results show that it outperforms methods built on lightweight open-source VLMs of comparable scale, as well as solutions based on closed-source VLMs.
\end{itemize}

\section{Related Work}
\label{sec:related}
\noindent\textbf{Training-based KB-VQA.}
Training-based methods optimize backbone or auxiliary parameters to associate visual evidence with task-relevant knowledge. Early approaches such as KAT combine implicit language-model knowledge with retrieved explicit knowledge, providing complementary sources for KB-VQA~\citep{gui2022kat}. MuKEA organizes visual objects, implicit relations, and factual answers into multimodal triplets, accumulating retrievable knowledge for visual--fact reasoning~\citep{ding2022mukea}. REVIVE incorporates object regions and their relations into knowledge retrieval and answer prediction, strengthening the use of local visual evidence in KB-VQA~\citep{lin2022revive}.
Subsequent work broadens knowledge integration: REVEAL jointly learns memory encoding, retrieval, and generation to incorporate image--text, question--answer, and knowledge-graph information into KB-VQA~\citep{hu2023reveal}. Later VLM-based methods train auxiliary components around frozen model backbones. MR-MKG trains multimodal knowledge encoders and cross-modal alignment modules to make visual and semantic graph relations accessible to language models~\citep{lee2024mrmkg}. QACap trains a question-aware captioner that translates answer-relevant visual details into text for a frozen language model~\citep{yang2025qacap}. These developments improve knowledge access and visual information transfer, but freezing backbones does not remove the need for additional parameter optimization and suitable training data.

\textbf{Training-free KB-VQA.}
Training-free methods combine frozen pretrained models through prompting, retrieval, and tool use, without additional parameter training or adaptation of any component. Early approaches such as PICa use captions, tags, and example question--answer pairs to invoke language-model knowledge for KB-VQA~\citep{yang2022pica}. PnP-VQA exposes question-relevant visual details through local captions guided by the input question~\citep{tiong2022pnpvqa}. Img2LLM constructs captions and example question--answer pairs from images, providing both visual context and task demonstrations~\citep{guo2023img2llm}. Their effectiveness depends on preserving answer-relevant visual details in the textual context.
Subsequent approaches integrate external information acquisition with reasoning to supplement parametric knowledge. AVIS dynamically selects visual tools and retrieval operations based on intermediate results, allowing KB-VQA systems to actively seek missing information~\citep{hu2023avis}. MG$^2$-RAG organizes textual entities, visual regions, and document chunks into a multi-granularity graph, retrieving structured evidence for visual questions requiring multiple knowledge connections~\citep{dai2026mg2rag}. However, retrieval relevance alone does not establish evidential support for an answer. Retrieved information inconsistent with the image, or conflicting visual observations, can still interfere with reasoning. Assessing the validity and consistency of visual evidence alongside knowledge acquisition therefore remains an important challenge in KB-VQA.
Beyond individual queries, early memory-augmented agents retain feedback or skills for subsequent reuse. Reflexion uses reflections on previous attempts to inform later decisions, while Voyager reuses acquired capabilities through a library of executable skills~\citep{shinn2023reflexion,wang2023voyager}. Later, A-MEM dynamically links and revises historical memories; Mem$^p$ retrieves trajectories and abstract procedural steps as reasoning context and refines memories through feedback~\citep{xu2025amem,fang2026memp}. These general-purpose agents offer mechanisms that can inform experience accumulation and reuse in KB-VQA. However, semantic similarity to a previous problem does not guarantee that its knowledge or procedures apply to a new image. Aligning experience reuse with current visual evidence therefore remains an open question. In contrast, \method grounds both knowledge retrieval and memory reuse in co-verified visual evidence and refines experience through feedback.

\section{Method}
\label{sec:method}

\subsection{Framework Overview}

\begin{figure*}[t]
\centering
\includegraphics[width=\linewidth]{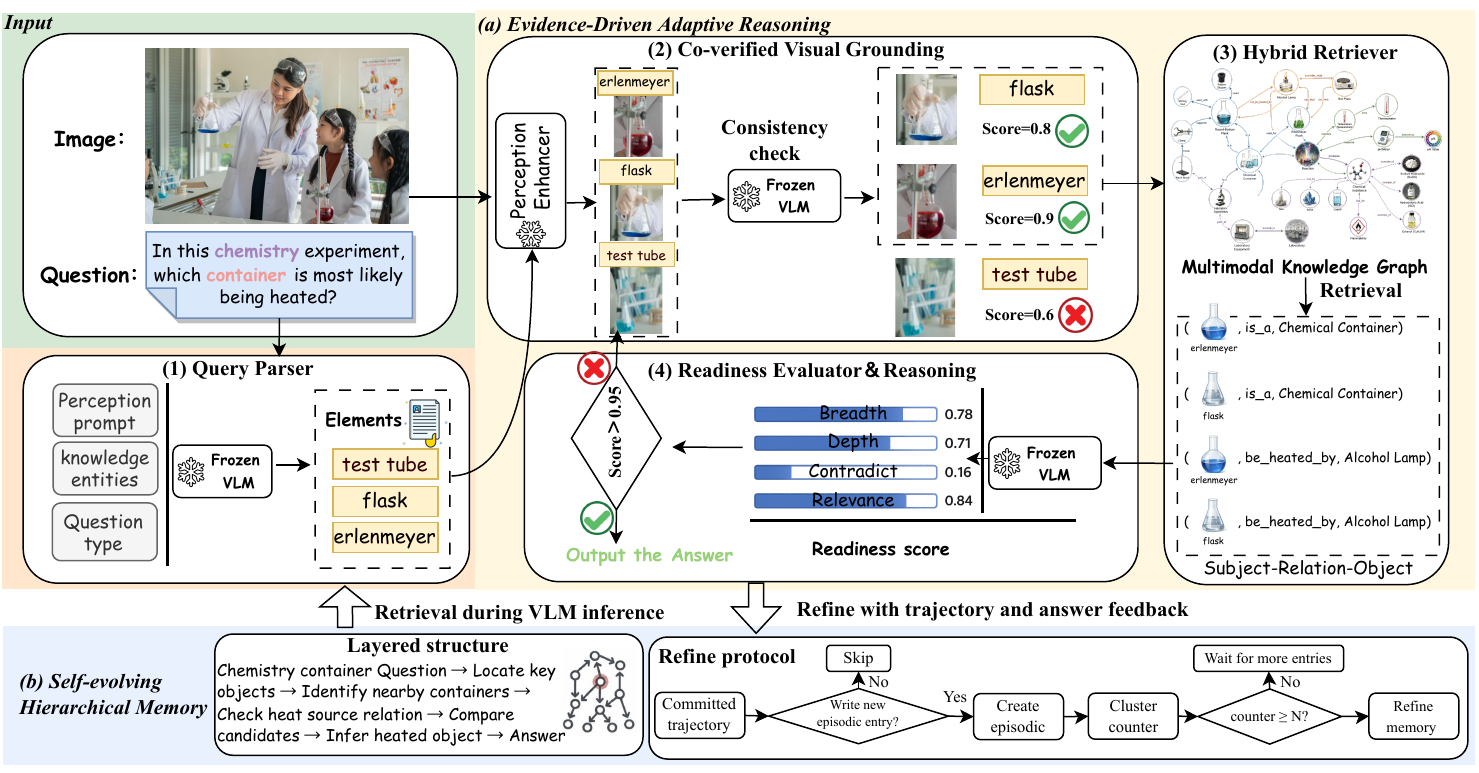}
\caption{Overview of \method. The framework comprises two core modules: Co-verified Visual Grounding for verifying question-relevant visual targets, and SHM for feedback-driven memory construction and refinement. EDAR integrates visual evidence, external knowledge, and relevant memories, then uses readiness assessment to guide further evidence gathering or answer generation.}
\label{fig:framework}
\end{figure*}

SAGE-VQA is a training-free agentic framework that combines evidence verification with experience evolution. Figure~\ref{fig:framework} illustrates the framework. It uses two modules: Co-verified Visual Grounding and Self-evolving Hierarchical Memory (SHM). Co-verified Visual Grounding locates and verifies question-relevant visual targets. SHM builds and refines a memory bank from post-answer feedback for use on later questions. The Evidence-Driven Adaptive Reasoning (EDAR) strategy acquires and verifies visual evidence, reasons with external knowledge and relevant memories, and uses readiness assessment to decide whether to seek further evidence or answer. Task notation and core prompt templates appear in Appendices~\ref{app:task-formulation} and~\ref{app:core-prompts}, respectively.

\subsection{Co-verified Visual Grounding}
\label{sec:method-grounding}

\textbf{Insight.}
A model may fail to identify question-relevant objects, attributes, or relations in an image. Subsequent reasoning can then rely on incorrect visual information and produce an unreliable answer. These errors can also mislead external knowledge retrieval. We therefore design Co-verified Visual Grounding to locate and verify local candidates before using visual information for reasoning.

\textbf{Open-vocabulary visual prompts.}
A frozen VLM serves as the fact reasoner $\mathcal M$. At iteration $t$, $\mathcal M$ provides a set of open-vocabulary visual prompts $\mathcal P_t$. Each prompt describes a target object in text and guides localization in the image. Below, task subscripts on $f$ distinguish calls to the same frozen model $\mathcal M$ for different tasks.

\textbf{Co-verified Visual Grounding.}
A pretrained segmentation model serves as the perception enhancer $\mathcal S_{\mathrm{seg}}$. It uses open-vocabulary visual prompts to locate and segment candidate regions in image $I$. The reasoner $\mathcal M$ checks whether these regions match the concepts in the prompts to assess semantic consistency. Let $p_{t,j}$ denote a prompt used at iteration $t$. Let $m_{t,j,k}$ and $b_{t,j,k}$ denote the mask and bounding box of its $k$-th candidate. We verify each candidate using its masked crop and visual prompt:
\begin{equation}
(a_{t,j,k},v_{t,j,k})
=f_{\mathrm v}\!\left(I[m_{t,j,k}],p_{t,j}\right),
\label{eq:verify}
\end{equation}
where $f_{\mathrm v}$ denotes visual candidate verification. The index $t\in\{1,\ldots,T\}$ denotes the iteration within one question, and $T$ is the maximum number of iterations. The subscript $t$ identifies the prompts, candidate regions, and verification results used at iteration $t$. Unchanged information is reused without repeating segmentation or verification. The index $j$ identifies a prompt, and $k$ identifies a candidate for that prompt. The term $I[m_{t,j,k}]$ denotes the masked crop. The value $a_{t,j,k}\in\{0,1\}$ records the semantic match, and $v_{t,j,k}\in[0,1]$ is the model-reported confidence.

The system retains the candidates and their verification annotations as $\mathcal A_{t,j}=\{(m_{t,j,k},b_{t,j,k},a_{t,j,k},v_{t,j,k})\}_{k=1}^{K_{t,j}}$, where $K_{t,j}$ is the number of candidates. The reasoner $\mathcal M$ then compares the candidates with its whole-image observations. It checks for disagreements in objects, counts, and relations. Candidate records and whole-image observations jointly form the visual evidence $\mathcal V_t$. Failed candidates retain their failure annotations so the reasoner can assess their impact. Appendix~\ref{app:visual-grounding} details candidate verification, whole-image comparison, and visual rechecking.

\subsection{Self-evolving Hierarchical Memory Mechanism}
\label{sec:method-memory}

\begin{figure*}[t]
\centering
\makebox[\linewidth][c]{%
\hspace*{-5mm}%
\includegraphics[width=0.8\linewidth,
  trim=0 202bp 0 65bp,
  clip]{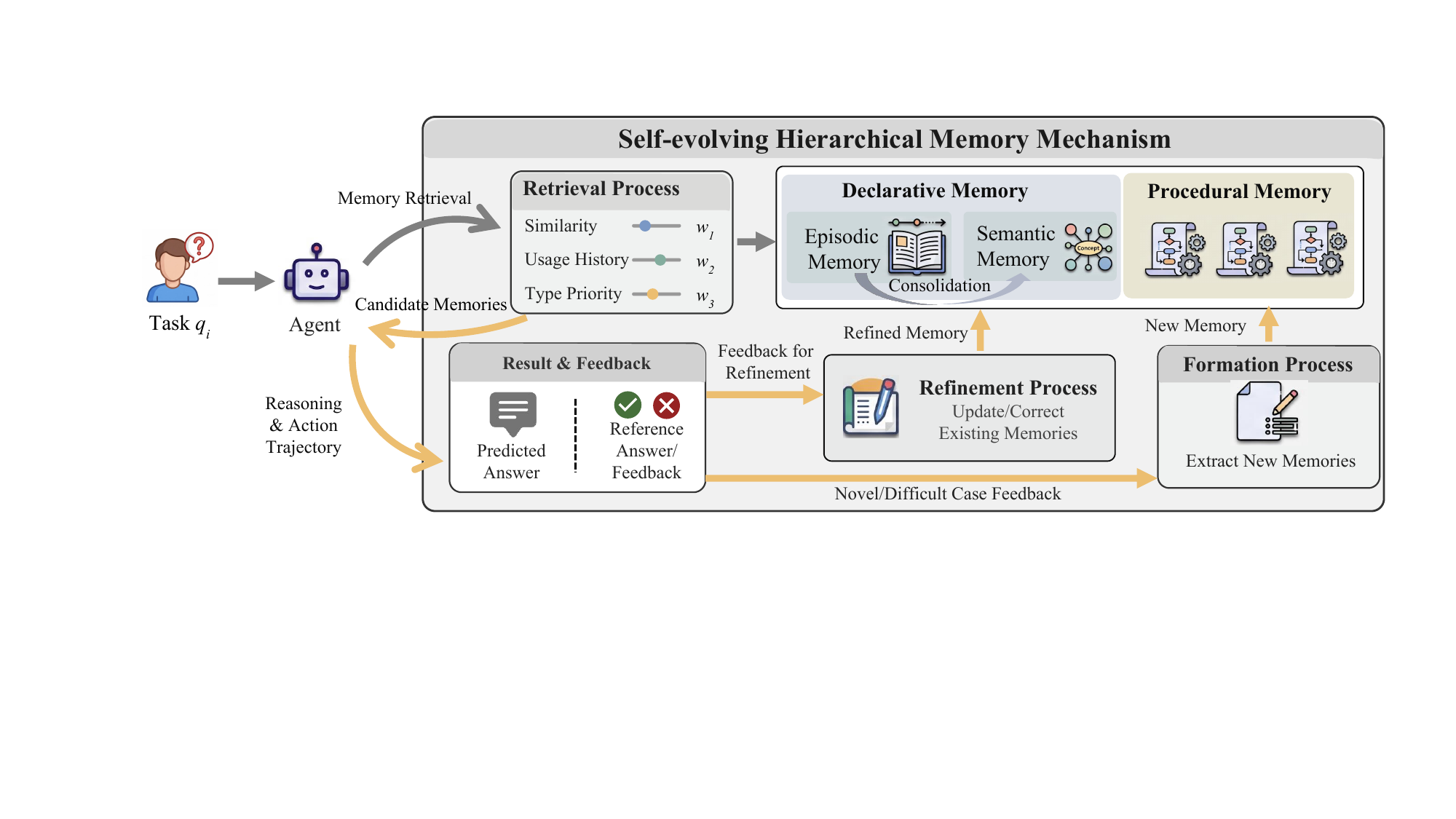}%
\hspace*{5mm}%
}
\caption{Self-evolving Hierarchical Memory (SHM). SHM organizes episodic records and semantic rules as declarative memory, and executable methods as procedural memory. Based on feedback, SHM can refine existing memories or add new memory.}
\label{fig:memory}
\end{figure*}

\textbf{Memory organization and retrieval.}
Inspired by the episodic--semantic distinction~\citep{tulving1972episodic} and memory organization in language agents~\citep{sumers2024coala}, SHM organizes its bank into declarative and procedural memory (Figure~\ref{fig:memory}). Declarative memory combines episodic memory of task-specific reasoning and corrections with generalizable semantic memory distilled from related episodes. Procedural memory stores executable solution methods and their applicability conditions. The initial bank is built from training examples disjoint from evaluation using reasoning trajectories and post-answer feedback from reference answers, with model parameters fixed. For each question, SHM retrieves relevant entries from the latest completed bank based on similarity to memory titles, usage history, and memory type. Appendices~\ref{app:memory-examples} and~\ref{app:memory-formation} detail the memory representations, cold-start construction, and retrieval procedure.

\textbf{Memory refinement.}
After answering a question, the system receives feedback based on the reference answer $y_i$. Let $\tau_i$ be the actual number of iterations and $\Pi_i$ the complete reasoning record. The current record contains $Q_i$, $\Pi_i$, the prediction $\hat A_i$, the reference answer, and correctness $c_i\in\{0,1\}$ under the task protocol. A correct answer reinforces the usage statistics of relevant memories. An incorrect answer triggers refinement of entries that remain applicable but need revision:
\begin{equation}
\mu_\ell^{\prime}
=f_{\mathrm m}
\!\left(\mu_\ell,Q_i,\Pi_i,\hat A_i,y_i\right),
\label{eq:memory-refine}
\end{equation}
where $f_{\mathrm m}$ refines an existing memory using the current case and feedback. The term $\mu_\ell^{\prime}$ denotes the refined entry. Refinement can correct knowledge descriptions, add applicability conditions, or revise program content. An inapplicable retrieved entry calls for an adjustment to its retrieval priority rather than a content rewrite based on that case. The original entry is retained if refinement produces no valid revision. This mechanism improves experience for similar questions without adding every reasoning record as a new memory. It therefore limits redundant growth of the bank.

\textbf{New memory formation and hierarchical organization.}
Let $\mathcal U_i$ denote the memories selected for the current question. When this set is empty, new experience can be extracted from incorrect answers or correct answers requiring multiple iterations:
\begin{equation}
\eta_i=\mathbf 1
\!\left[\mathcal U_i=\varnothing\ \land\
\bigl(c_i=0\ \lor\ (c_i=1\land\tau_i>1)\bigr)\right],
\label{eq:memory-admission}
\end{equation}
where $\mathbf 1[\cdot]$ is the indicator function. The condition $\eta_i=1$ permits extraction; only valid content is inserted. Episodes sharing a capability gap and similar solution patterns can form semantic memory once the group meets the size requirement. Recurring computational needs can support the formation or refinement of procedural memory. The bank update combines memory replacement, insertion, and hierarchical organization:
\begin{equation}
\mathcal B^{(s+1)}=h\bigl(\{\mu\in\mathcal B^{(s)}:\mu\notin\mathcal F_i\}\cup\mathcal F_i^{\prime}\cup\mathcal N_i\bigr),
\label{eq:memory-operation}
\end{equation}
where $h$ organizes the memory set into the hierarchy under the conditions above. $\mathcal B^{(s)}$ denotes the current write version, $\mathcal F_i$ the entries to replace, $\mathcal F_i^{\prime}$ their valid revisions, and $\mathcal N_i$ valid new entries. See supplementary Algorithm~\ref{alg:memory-update} for the complete memory update procedure.

\begin{table*}[t]
\caption{ScienceQA test-set comparison. TF: training-free across all components. Subject: NAT/SOC/LAN denote natural/social/language science; context modality: TXT/IMG/NO denote text/image/no context; grade: G1--6/G7--12; Avg.: average accuracy.}
\label{tab:sota-sciqa}
\centering
\small
\setlength{\tabcolsep}{2.5pt}
\renewcommand{\arraystretch}{1.05}
\begin{tabular}{@{}lcccccccccc@{}}
\toprule
\multirow{2}{*}{Method} & \multirow{2}{*}{TF}
& \multicolumn{3}{c}{Subject} & \multicolumn{3}{c}{Context Modality}
& \multicolumn{2}{c}{Grade} & \multirow{2}{*}{Avg.} \\
\cmidrule(lr){3-5} \cmidrule(lr){6-8} \cmidrule(lr){9-10}
& & NAT & SOC & LAN & TXT & IMG & NO & G1-6 & G7-12 & \\
\midrule
Human~\citep{lu2022scienceqa}
  & - & 90.23 & 84.97 & 87.48 & 89.60 & 87.50 & 88.10 & 91.59 & 82.42 & 88.40 \\
\midrule
Multimodal-CoT~\citep{zhang2024mmcot}
  & \ding{55} & 91.03 & 93.70 & 86.64 & 90.13 & 88.25 & 89.48 & 91.12 & 89.26 & 90.45 \\
LLaVA~\citep{liu2023llava}
  & \ding{55} & 90.36 & 95.95 & 88.00 & 89.49 & 88.00 & 90.66 & 90.93 & 90.90 & 90.92 \\
LLaVA+GPT-4 judge~\citep{liu2023llava}
  & \ding{55} & 91.56 & 96.74 & 91.09 & 90.62 & 88.99 & 93.52 & 92.73 & 92.16 & 92.53 \\
CalibRL (Qwen2.5-VL-7B)~\citep{calibrl2026}
  & \ding{55} & - & - & - & - & - & - & - & - & 89.04 \\
LLaMA-Adapter~\citep{zhang2024llamaadapter}
  & \ding{55} & 84.37 & 88.30 & 84.36 & 83.72 & 80.32 & 86.90 & 85.83 & 84.05 & 85.19 \\
LaVIN-13B~\citep{luo2023lavin}
  & \ding{55} & 89.88 & 94.49 & 89.82 & 88.95 & 87.61 & 91.85 & 91.45 & 89.72 & 90.83 \\
MR-MKG (FLAN-UL2-19B)~\citep{lee2024mrmkg}
  & \ding{55} & 95.74 & 90.33 & 92.00 & 95.50 & 92.41 & 93.31 & 93.98 & 93.01 & 93.63 \\
RG-VQA (Bunny-8B)~\citep{sravanthi2025rgvqa}
  & \ding{55} & 90.45 & \underline{96.85} & 91.64 & 95.37 & 88.50 & 93.24 & 92.47 & 91.43 & 92.10 \\
\midrule
MG$^2$-RAG(Qwen3.5-27B)~\citep{dai2026mg2rag}
  & \checkmark & 98.45 & 95.28 & 98.73 & \underline{98.44} & 96.43 & 98.68 & 98.42 & 96.84 & 97.85 \\
Gemini-3.1-Pro~\citep{dai2026mg2rag}
  & \checkmark &93.53 & 94.12 & 90.81 & 92.98 & 94.25 & 90.27 & 94.03 & 90.93 & 92.90 \\
GPT-5.2~\citep{dai2026mg2rag}
  & \checkmark & 97.60 & 93.05 & 95.96 & 97.73 & 95.35 & 96.17 & 96.54 & 95.88 & 96.30 \\
\midrule
\textbf{\method (GPT-4o-mini)}
  & \checkmark & 88.90 & 85.94 & 90.00 & 86.95 & 83.94 & 91.64 & 90.86 & 84.44 & 88.56 \\
\textbf{\method (Qwen3.5-9B)}
  & \checkmark & 95.29 & 91.79 & 98.27 & 94.23 & 91.57 & 98.40 & 96.84 & 92.62 & 95.33 \\
\textbf{\method (Qwen3.5-27B)}
  & \checkmark & \underline{98.67} & \underline{96.85} & \textbf{99.73} & \textbf{98.68} & \underline{97.47} & \textbf{99.51} & \underline{98.97} & \underline{97.82} & \underline{98.56} \\
\textbf{\method (Qwen3.6-35B)}
  & \checkmark & \textbf{99.02} & \textbf{97.86} & \underline{99.55} & \textbf{98.68} & \textbf{98.26} & \underline{98.75} & \textbf{99.27} & \textbf{98.29} & \textbf{98.92} \\
\bottomrule
\end{tabular}
\end{table*}

\subsection{Evidence-Driven Adaptive Reasoning Strategy}
\label{sec:method-edar}

EDAR integrates visual evidence, external knowledge, and relevant memories, using evidence sufficiency to decide whether to seek more evidence or answer. After recording the answer, SAGE-VQA passes the completed case to SHM if feedback updates are enabled and a reference answer is available. Algorithm~\ref{alg:system} outlines the process; Appendix~\ref{app:reasoning-procedure} provides further details.

\textbf{Initialization and multimodal evidence acquisition.}
Given image $I$ and question $Q$, $\mathcal M$ obtains an open-vocabulary visual prompt set $\mathcal P_1$ and a knowledge entity set $\mathcal E_1$ from the question and global image description $C$. We write $(\mathcal P_1,\mathcal E_1)=f_{\mathrm p}(Q,C)$, where $f_{\mathrm p}$ denotes question parsing. The prompts are passed to Co-verified Visual Grounding in Section~\ref{sec:method-grounding} to form visual evidence $\mathcal V_t$. The entities are used to query the multimodal knowledge graph. The system retrieves relevant memories $\mathcal U_Q$ using Section~\ref{sec:method-memory}. Procedural memory executes when its input requirements are met. Inputs may come from the question, image descriptions, or acquired visual information. The set $\mathcal W_t$ contains the validated outputs of executed procedures.

The system queries the multimodal knowledge graph $\mathcal G$ using knowledge entities. It also uses locally verified image regions for visual retrieval. Let $\mathcal E_t$ and $\mathcal O_t$ denote the entities and image regions used for retrieval at iteration $t$. The knowledge update is:
\begin{equation}
\mathcal K_t
=\mathcal K_{t-1}\cup
g\!\left(\mathcal E_t,\mathcal O_t\right),
\label{eq:knowledge-update}
\end{equation}
where $g$ denotes retrieval from the $\mathcal G$ using entities and image regions. The set $\mathcal K_t$ contains the graph knowledge acquired through iteration $t$. Entity retrieval returns relevant relations and facts. Visual retrieval locates graph entities through similar images. The initial retrieval uses both channels. Later queries target missing information. Appendix~\ref{app:knowledge-retrieval} describes both retrieval channels.

\textbf{Readiness assessment and answering.}
$\mathcal M$ assesses whether the visual evidence, graph knowledge, relevant memories, validated program outputs, and reasoning record are sufficient to answer the question. It assigns a readiness score between 0 and 1 and identifies unresolved issues in breadth, depth, contradiction, or relevance. If readiness remains below the threshold and iterations remain, $\mathcal M$ uses these issues and the reasoning record to select visual rechecking or knowledge supplementation. Conflicting or missing visual details trigger focused visual rechecking, whereas missing facts or definitions trigger knowledge retrieval. During visual rechecking, existing segmentation results can serve as prompts for the reasoner to inspect unresolved image details. The system reassesses readiness after each iteration and answers using this information once readiness reaches the threshold or the iteration budget is exhausted. Appendix~\ref{app:readiness} details issue categories and stopping criteria.

\textbf{Inference cost.}
Memory retrieval runs once per question, and each iteration performs one readiness assessment. Grounding uses $O(mK)$ verification calls for $m$ prompts with at most $K$ candidates each, and later iterations reuse unchanged results. After returning the answer, the system updates the memory bank asynchronously.

\section{Experiments}
\label{sec:exp}

\subsection{Setup}
We evaluate four settings: ScienceQA~\citep{lu2022scienceqa} on its official test split; all 1{,}145 A-OKVQA~\citep{schwenk2022aokvqa} validation questions in both answer formats; all 900 MMMU~\citep{yue2024mmmu} validation questions; and 1{,}000 validation questions from MicroQuant-VQA, a private dataset we constructed(appendix~\ref{app:microquant-example}). We use top-1 accuracy to evaluate multiple-choice questions and VQAv2 soft accuracy to evaluate direct-answer tasks. Appendix~\ref{app:datasets} provides full statistics.

\begin{table*}[t]
\centering
\begin{minipage}[t]{0.58\linewidth}
\raggedright
\caption{A-OKVQA validation. MC/DA: multiple choice/direct answer. QACap's occurrence-based DA score is omitted as non-comparable.}
\label{tab:sota-aokvqa}
\centering
\small
\setlength{\tabcolsep}{2.2pt}
\renewcommand{\arraystretch}{1.02}
\begin{tabular}{@{}lccc@{}}
\toprule
Method & TF & MC & DA \\
\midrule
ViLBERT~\citep{schwenk2022aokvqa} & \ding{55} & 49.1 & 30.6 \\
GPV-2~\citep{schwenk2022aokvqa} & \ding{55} & 60.3 & 48.6 \\
MAGIC-VQA~\citep{yang2025magicvqa} & \ding{55} & 76.55 & - \\
QACap+Claude-3.5~\citep{yang2025qacap} & \ding{55} & 76.7 & - \\
V3Fusion~\citep{tekin2025v3fusion} & \ding{55} & 89.17 & - \\
PromptCap+GPT-3~\citep{hu2023promptcap} & \ding{55} & 73.2 & 56.3 \\
\midrule
\textbf{\method (GPT-4o-mini)} & \checkmark & 80.26 & 61.46 \\
\textbf{\method (Qwen3.5-9B)} & \checkmark & 89.98 & 68.91 \\
\textbf{\method (Qwen3.5-27B)} & \checkmark & 90.82 & 69.84 \\
\textbf{\method (Qwen3.6-35B)} & \checkmark & \textbf{91.54} & \textbf{72.67} \\
\bottomrule
\end{tabular}
\end{minipage}\hfill
\begin{minipage}[t]{0.39\linewidth}
\raggedright
\caption{MMMU validation under the official mixed-format protocol.}
\label{tab:sota-mmmu}
\centering
\small
\setlength{\tabcolsep}{2.2pt}
\renewcommand{\arraystretch}{1.02}
\begin{tabular}{@{}lcc@{}}
\toprule
Method & TF & Acc. \\
\midrule
GPT-4V~\citep{yue2024mmmu} & \checkmark & 56.80 \\
GPT-4o-mini~\citep{openai2024gpt4omini} & \checkmark & 59.40 \\
Qwen2.5-VL-7B~\citep{bai2025qwen25vl} & \checkmark & 58.60 \\
\midrule
MAGIC-VQA~\citep{yang2025magicvqa} & \ding{55} & 61.03 \\
V3Fusion~\citep{tekin2025v3fusion} & \ding{55} & 56.09 \\
CalibRL~\citep{calibrl2026} & \ding{55} & 56.55 \\
\midrule
\textbf{\method (GPT-4o-mini)} & \checkmark & 64.59 \\
\textbf{\method (Qwen3.5-9B)} & \checkmark & 66.30 \\
\textbf{\method (Qwen3.5-27B)} & \checkmark & 75.32 \\
\textbf{\method (Qwen3.6-35B)} & \checkmark & \textbf{76.55} \\
\bottomrule
\end{tabular}
\end{minipage}
\par
\begin{minipage}[t]{0.49\linewidth}
\caption{Accuracy (\%) on 1{,}000 MicroQuant-VQA validation questions. Parentheses indicate gains in percentage points over the corresponding bare VLM.}
\label{tab:microquant-transfer}
\centering
\small
\begin{tabular}{@{}lc@{}}
\toprule
Configuration & Accuracy \\
\midrule
Bare VLM (9B) & $23.10$ \\
\textbf{\method (9B)} & $35.80$ ($+12.70$) \\
\midrule
Bare VLM (27B) & $29.30$ \\
\textbf{\method (27B)} & $\mathbf{87.50}$ ($+58.20$) \\
\bottomrule
\end{tabular}
\end{minipage}\hfill
\begin{minipage}[t]{0.49\linewidth}
\caption{Incremental ablation on Qwen3.5-9B. The EDAR row adds visual verification and adaptive reasoning to MMKG retrieval. Parentheses report gains over the preceding row.}
\label{tab:ablation}
\centering
\small
\setlength{\tabcolsep}{1.6pt}
\renewcommand{\arraystretch}{1.04}
\begin{tabular}{@{}lcc@{}}
\toprule
Configuration & ScienceQA & MMMU \\
\midrule
Bare VLM & $70.79$ & $50.12$ \\
\quad + MMKG & $76.96$ ($+6.17$) & $54.44$ ($+4.32$) \\
\quad + EDAR & $92.76$ ($+15.80$) & $61.29$ ($+6.85$) \\
\quad + SHM & $\mathbf{95.33}$ ($+2.57$) & $\mathbf{66.30}$ ($+5.01$) \\
\bottomrule
\end{tabular}
\end{minipage}
\end{table*}

We use Qwen3.5-9B as the main backbone and SAM3~\citep{carion2026sam3} as the perception enhancer. We also evaluate Qwen3.5-27B, Qwen3.6-35B, and GPT-4o-mini. The memory bank is initialized from data disjoint from the evaluation set. It continues to evolve during evaluation. Later questions use the latest completed memory version. Model parameters remain fixed. Further implementation details are provided in Appendices~\ref{app:hyperparams} and~\ref{app:baselines}.
\begin{algorithm}[t]
\caption{EDAR inference with SHM feedback.}
\label{alg:system}
\small
\begin{algorithmic}[1]
\Require $I$, $Q$, memory bank $\mathcal B^{(s_i)}$, threshold $\theta_{\mathrm r}$, budget $T$
\Ensure Answer $\hat A$, reasoning record $\Pi$
\State $C\gets$ caption of $I$; $(\mathcal P_1,\mathcal E_1)\gets f_{\mathrm p}(Q,C)$
\State Retrieve $\mathcal U_Q$ from $\mathcal B^{(s_i)}$; $\mathcal K_0,\mathcal H_0\gets\varnothing$
\For{$t=1$ \textbf{to} $T$}
  \State $\mathcal V_t\gets$ co-verified grounding of $\mathcal P_t$ \Comment{Eq.~\eqref{eq:verify}}
  \State $\mathcal K_t\gets\mathcal K_{t-1}\cup g(\mathcal E_t,\mathcal O_t)$ \Comment{Eq.~\eqref{eq:knowledge-update}}
  \State $\mathcal W_t\gets$ validated outputs of procedures in $\mathcal U_Q$
  \State $\mathcal H_t\gets$ reason over $\mathcal V_t,\mathcal K_t,\mathcal U_Q,\mathcal W_t,\mathcal H_{t-1}$
  \State $(r_t,\mathcal D_t)\gets$ readiness score and unresolved issues
  \State \textbf{if} $r_t\ge\theta_{\mathrm r}$ \textbf{or} $t=T$ \textbf{then} $\tau\gets t$; \textbf{break}
  \State Set $\mathcal P_{t+1}$ or $\mathcal E_{t+1}$ from $\mathcal D_t$ \Comment{recheck / supplement}
\EndFor
\State Generate $\hat A$ and finalize $\Pi$
\State \textbf{if} $y$ is available: update SHM asynchronously (Alg.~\ref{alg:memory-update})
\State \Return $(\hat A,\Pi)$
\end{algorithmic}
\end{algorithm}

\subsection{Comparison Results}
\label{sec:exp-sota}

\textbf{Overall results.}
Tables~\ref{tab:sota-sciqa}--\ref{tab:sota-mmmu} compare \method with training-based and training-free methods. With a lightweight Qwen3.5-9B backbone, \method outperformed MR-MKG on ScienceQA by $1.70$ percentage points. MR-MKG uses a frozen FLAN-UL2-19B backbone and requires task-specific training of auxiliary modules. On A-OKVQA multiple choice, our 9B configuration exceeded V3Fusion by $0.81$ percentage points. V3Fusion trains an MLP to fuse a three-model ensemble that includes 13B and 7B LLaVA backbones. On MMMU, our 9B configuration surpassed MAGIC-VQA with GPT-4o-mini by $5.27$ percentage points. MAGIC-VQA additionally trains a lightweight GNN for commonsense reasoning. The same 9B configuration also exceeded Gemini-3.1-Pro on ScienceQA and QACap+Claude-3.5 on A-OKVQA multiple choice by $2.43$ and $13.28$ percentage points, respectively. On A-OKVQA direct answers, it surpassed PromptCap+GPT-3 by $12.61$ percentage points. PromptCap combines a trained captioner with GPT-3. The 27B configuration further outperformed GPT-5.2 on ScienceQA by $2.26$ percentage points.
These comparisons show that \method can outperform methods requiring auxiliary-component training or closed-source models while using a lightweight open-source reasoning backbone. The direct-answer gains extend this advantage beyond selecting from predefined options.

\textbf{MicroQuant-VQA.}\label{sec:exp-microquant}
We further evaluated \method on MicroQuant-VQA, which requires quantitative measurements from micro-CT images. As shown in Table~\ref{tab:microquant-transfer}, the 9B and 27B configurations achieved $35.80\%$ and $87.50\%$, improving over their bare backbones by $12.70$ and $58.20$ percentage points. These results support the framework's applicability to materials analysis through visual grounding and reusable computational procedures. Appendix~\ref{app:microquant-example} provides the task details.

\begin{table}[ht]
\caption{Feedback-driven memory self-evolution on 84 recurring ScienceQA hard cases. $\Delta$ is measured from Pass~1; transitions are relative to the preceding pass.}
\label{tab:online-evolution}
\centering
\small
\setlength{\tabcolsep}{2.2pt}
\renewcommand{\arraystretch}{1.06}
\begin{tabular}{@{}cccccc@{}}
\toprule
Pass & Correct & Acc. (\%) & $\Delta$ (pp) & Corrected & Regressed \\
\midrule
1 & 32 & 38.10 & -- & -- & -- \\
2 & 38 & 45.24 & +7.14 & 15 & 9 \\
3 & 40 & 47.62 & +9.52 & 14 & 12 \\
4 & 41 & \textbf{48.81} & \textbf{+10.71} & 11 & 10 \\
5 & 40 & 47.62 & +9.52 & 9 & 10 \\
\bottomrule
\end{tabular}
\end{table}

\begingroup
\begin{table*}[t]
\caption{Representative error recovery.}
\label{tab:casestudy}
\centering
\small
\setlength{\tabcolsep}{3.4pt}
\renewcommand{\arraystretch}{1.0}
\begin{tabular}{@{}p{0.26\linewidth} p{0.23\linewidth} p{0.29\linewidth} p{0.14\linewidth}@{}}
\toprule
\textbf{Question (gold)} & \textbf{Baseline/prior state (wrong)} & \textbf{\method (correct)} & {\raggedright\textbf{Effective component}\par} \\
\midrule
\begin{minipage}[t]{0.50\linewidth}
\vspace{0pt}\includegraphics[width=\linewidth]{images/aokvqa_knicks_481390.jpg}
\end{minipage}\hfill
\begin{minipage}[t]{0.46\linewidth}
\vspace{0pt}\raggedright
Which team is in white/orange? (\emph{Knicks})
\end{minipage}
& \vspace{0pt}The bare VLM fixates on the opposing dark jersey (``JACKSON,'' No.~1), never establishes the white/orange team's identity, and predicts \textcolor{red}{Raptors}.
& \vspace{0pt}Although the VLM misreads a jersey as ``UTAH,'' MMKG retrieval links the image to the New York Knicks. The fact reasoner checks this identity against the choices and orange-white color scheme, yielding \textcolor{teal}{Knicks}.
& \vspace{0pt}{\raggedright \textbf{MMKG retrieval} + evidence arbitration.\par} \\
\midrule
\begin{minipage}[t]{0.50\linewidth}
\vspace{0pt}\includegraphics[width=\linewidth]{images/aokvqa_arms_384468.jpg}
\end{minipage}\hfill
\begin{minipage}[t]{0.46\linewidth}
\vspace{0pt}\raggedright
How many people have their arms around the woman? (\emph{one})
\end{minipage}
& \vspace{0pt}It conflates the three riders with the relation-specific count and predicts \textcolor{red}{three}, even though its own trace observes that only the man behind the woman has an arm around her.
& \vspace{0pt}The VLM separates group membership from the queried relation: the rear man embraces the woman, whereas the front rider does not. SAM3 independently returns one relevant detection, supporting \textcolor{teal}{one}.
& \vspace{0pt}{\raggedright \textbf{EDAR} enforces relation-aware counting.\par} \\
\midrule
\begin{minipage}[t]{0.50\linewidth}
\vspace{0pt}
\resizebox{\linewidth}{!}{%
\begin{tikzpicture}[x=0.88cm,y=0.44cm,font=\fontsize{6}{6.5}\selectfont]
\foreach \y/\name in {1.25/Pair 1,-0.45/Pair 2}{
  \node at (1.2,{\y+0.48}) {\name};
  \fill[green!60!black] (0,{\y-0.30}) rectangle (0.375,{\y+0.30});
  \fill[blue!65] (0.375,{\y-0.30}) rectangle (0.75,{\y+0.30});
  \fill[green!60!black] (1.65,{\y-0.30}) rectangle (2.025,{\y+0.30});
  \fill[blue!65] (2.025,{\y-0.30}) rectangle (2.40,{\y+0.30});
  \draw (0,{\y-0.30}) rectangle (0.75,{\y+0.30});
  \draw (1.65,{\y-0.30}) rectangle (2.40,{\y+0.30});
  \node[white] at (0.19,\y) {S}; \node[white] at (0.56,\y) {N};
  \node[white] at (1.84,\y) {S}; \node[white] at (2.21,\y) {N};
  \draw[<->] (0.75,{\y-0.43}) -- node[below] {1.75 in} (1.65,{\y-0.43});
}
\end{tikzpicture}}
\end{minipage}\hfill
\begin{minipage}[t]{0.46\linewidth}
\vspace{0pt}\raggedright
Which pair has the greater magnetic-force magnitude? (\emph{the same})
\end{minipage}
& \vspace{0pt}An earlier Qwen3.6-35B pass overweights a local cue and predicts \textcolor{red}{Pair 1}, despite identical materials, orientations, and separations.
& \vspace{0pt}Procedural memory invokes \texttt{compute\_}\allowbreak\texttt{magnetic\_}\allowbreak\texttt{force\_}\allowbreak\texttt{magnitude}. A successful execution compares the verified attributes and equal 1.75-in gaps, yielding \textcolor{teal}{the same}.
& \vspace{0pt}{\raggedright \textbf{Procedural-memory execution}\newline + verified measurement.\par} \\
\bottomrule
\end{tabular}
\end{table*}
\endgroup

\subsection{Ablation Study}
\label{sec:exp-ablation}

Table~\ref{tab:ablation} presents the ablation results using Qwen3.5-9B as the backbone. MMKG retrieval alone improved performance on both datasets. The EDAR row adds visual verification and adaptive reasoning to that retrieval baseline, thereby completing the EDAR strategy. This stage produced the largest incremental gain in the reported ordering. Adding SHM further improved performance through memory reuse and updating. The 27B ablation in Appendix~\ref{app:ablation-27b} showed the same pattern with a stronger backbone.

\subsection{Feedback-Driven Memory Evolution}
\label{sec:exp-online-evolution}

\noindent\textbf{Protocol.}
Using Qwen3.6-35B, we make five passes over 84 ScienceQA errors from an earlier full-system run, each inheriting the preceding bank. Feedback updates usage statistics and asynchronously refines applicable retrieved entries after errors. Cases without relevant retrieved memory can supply new experience after an error or a correct answer reached through multiple iterations. Only later queries can benefit; model parameters remain fixed.

\noindent\textbf{Result.}
As shown in Table~\ref{tab:online-evolution}, accuracy reached to a peak of $48.81\%$ in Pass~4. It then declined to $47.62\%$ in the final pass. Across passes, some previously incorrect cases were corrected, while some previously correct cases became incorrect. These results show improved overall accuracy alongside performance regressions, highlighting the need to assess and refine memories during continued use.

\subsection{Backbone Scaling Study}
\label{sec:exp-scaling}

\begin{figure}[t]
\centering
\includegraphics[width=\linewidth]{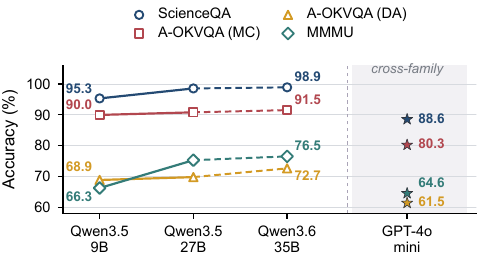}
\caption{Accuracy across fact-reasoner backbones. Lines connect Qwen backbones; stars denote GPT-4o-mini.}\label{fig:scaling}
\end{figure}

We evaluate \method with fact reasoners of different scales and families under common retrieval and iteration budgets (Figure~\ref{fig:scaling}). Within Qwen3.5, scaling from 9B to 27B improves accuracy on every benchmark and answer format, with the largest gain on MMMU ($+9.02$ points, from $66.30\%$ to $75.32\%$). The Qwen3.6-35B adds a further $0.36$--$2.83$ points, most notably on A-OKVQA direct answers. With GPT-4o-mini, \method reaches $64.59\%$ on MMMU, $5.19$ points above the reported GPT-4o-mini result in Table~\ref{tab:sota-mmmu}, showing that the framework transfers beyond the Qwen family. Overall, \method remains effective with lightweight backbones and benefits from stronger ones, indicating that backbone capability and the proposed reasoning framework are complementary.

\subsection{Case Studies}
\label{sec:exp-cases}

\noindent\textbf{Analysis.}
Table~\ref{tab:casestudy} shows that \method routes distinct capability gaps to complementary repairs. First, the bare VLM fixates on the opposing jersey; MMKG retrieval identifies the Knicks despite a misread jersey, and the fact reasoner confirms it against the color scheme. Second, the bare VLM counts all three riders although its own trace notes a single embrace; EDAR separates group membership from the queried relation, supported by one SAM3 detection. Third, an earlier pass overweights a local cue and predicts Pair~1, whereas procedural memory executes a force comparison over verified attributes. Each repair matches the diagnosed gap and is validated before answering.

\section{Conclusion}
\label{sec:conclusion}
\vspace{-1pt}

We propose \method, a training-free agentic framework for KB-VQA that combines evidence verification with experience evolution. Co-verified Visual Grounding verifies question-relevant visual information, while SHM organizes declarative and procedural memories and refines them through feedback. The EDAR strategy acquires and verifies visual evidence, retrieves external knowledge and relevant memories, and uses readiness assessment to guide further evidence acquisition or answering. Experiments across diverse benchmarks support the effectiveness of combining evidence verification with experience reuse. These gains extend across open-source and closed-source backbones without updating model parameters.

\clearpage
\appendix
\section*{Supplementary Material}
% Continue the source manuscript's figure, table, and equation numbering.
\setcounter{figure}{4}
\setcounter{table}{7}
\setcounter{equation}{5}
\setcounter{algorithm}{1}
\section{Supplementary Methodology}
\label{app:operational-details}

\subsection{Task Definition and Notation}
\label{app:task-formulation}

Given an image $I$ and question $Q$, KB-VQA combines visual information with external knowledge to produce an answer $\hat A$. The question input includes answer options and background text permitted by the evaluation protocol. For ScienceQA, this includes lecture annotations but excludes reference rationales. Reference-answer feedback is never part of the current question input. All model parameters remain fixed.

The frozen fact reasoner $\mathcal M$ first produces a global image description $C$. It parses this description and the question into open-vocabulary visual prompts and knowledge entities:
\begin{equation}
(\mathcal P_1,\mathcal E_1)=f_{\mathrm p}(Q,C).
\label{eq:parse}
\end{equation}
where, $\mathcal P_1=\{p_{1,j}\}_{j=1}^{m}$ contains textual descriptions of target objects. The set $\mathcal E_1$ contains entities for knowledge retrieval. Task subscripts on $f$ distinguish calls to the same frozen model: $f_{\mathrm p}$ parses the question, $f_{\mathrm v}$ verifies visual candidates, $f_{\mathrm r}$ assesses readiness, and $f_{\mathrm m}$ refines memories. Answer generation also uses the frozen model. These functions do not introduce separately trained models.

The index $t$ identifies an EDAR iteration within one question. The index $i$ identifies a question across the task stream. The actual stopping iteration is $\tau$ for a single question and $\tau_i$ for question $Q_i$. The sets $\mathcal V_t$, $\mathcal K_t$, and $\mathcal W_t$ contain visual evidence, accumulated graph knowledge, and validated program outputs. Graph knowledge starts empty before the first retrieval. The record $\mathcal H_{t-1}$ contains reasoning and actions from earlier iterations. The complete record $\Pi_i$ preserves the reasoning and actions through answer generation. It supports memory feedback only after the answer is produced.

The superscript $(s)$ identifies a completed memory version. Question $Q_i$ reads $\mathcal B^{(s_i)}$ and selects a fixed memory set $\mathcal U_i$. For the same question, $\mathcal U_Q$ denotes this set in the single-question reasoning procedure. A background update may begin from a later version $\mathcal B^{(s)}$. It does not change the memory context already selected for an ongoing question.

\subsection{Core Prompt Templates}
\label{app:core-prompts}

Table~\ref{tab:core-prompts} presents visual candidate verification, Table~\ref{tab:core-prompts-arbitration} covers evidence arbitration and action selection, and Table~\ref{tab:core-prompts-refinement} describes feedback-based memory refinement. These core prompt templates are condensed for presentation; braces denote question-specific inputs.

\begingroup
\setlength{\fboxsep}{6pt}
\setlength{\fboxrule}{0.6pt}
\newcommand{\corepromptbox}[2]{%
\fcolorbox{black!70}{black!3}{%
\begin{minipage}{\dimexpr\linewidth-2\fboxsep-2\fboxrule\relax}
\colorbox{black!75}{\parbox{\dimexpr\linewidth-2\fboxsep\relax}{\color{white}\bfseries #1}}\par\smallskip
\raggedright #2
\end{minipage}}}

\begin{table*}[t]
\caption{Visual candidate verification prompt.}
\label{tab:core-prompts}
\centering
\corepromptbox{Visual Candidate Verification}{%
Look at this image crop. Does it primarily show a `\{target object\}'?
Reply ONLY with a valid JSON block, nothing else.\par\smallskip
Format: \texttt{\{"is\_match": true, "confidence\_score": 0.9\}}.
}
\end{table*}

\begin{table*}[t]
\caption{Evidence arbitration and action selection prompt.}
\label{tab:core-prompts-arbitration}
\centering
\corepromptbox{Evidence Arbitration and Action Selection}{%
You are a Scientific Reasoning Adjudicator. Answer by weighing evidence from multiple agents.\par\smallskip
Question and choices: \{question, choices\}.\par
Evidence: \{visual observations, verified candidates, external knowledge, relevant memories\}.\par
Evaluator feedback and reasoning history: \{feedback, history\}.\par\smallskip
Task 1: Resolve disagreements between VLM observations and SAM3 detections. Check counts because SAM3 can miss objects. For abstract concepts and relationships, rely on the VLM.\par\smallskip
Task 2: Use evaluator feedback to choose among answering, focused visual rechecking, and further knowledge retrieval. Request visual rechecking for conflicting or missing visual details, and knowledge retrieval for missing facts or definitions.\par\smallskip
Task 3: When instructed to commit, provide the final answer. Return your answer or requested action with a brief justification.
}
\end{table*}

\begin{table*}[t]
\caption{Feedback-based memory refinement prompt.}
\label{tab:core-prompts-refinement}
\centering
\corepromptbox{Feedback-Based Memory Refinement}{%
You are the meta-cognitive reflection module. A previously stored memory has failed. Analyse and upgrade it.\par\smallskip
Existing memory: \{memory\}.\par
Failed case: \{question, choices, prediction, reference answer\}.\par
Original reasoning history: \{reasoning history\}.\par\smallskip
Task 1: Diagnose the failure as a strategy, computation, knowledge, or perception gap.\par\smallskip
Task 2: Explain what the old memory missed and revise its reusable guidance. The revised steps should subsume the previous advice.\par\smallskip
Task 3: For a computation gap, improve the existing executable procedure or replace a repeatedly failing textual strategy with a procedure.\par\smallskip
Generalise: avoid copying unique numbers or proper nouns from this case. Return a short title, a description of the correction, and the revised guidance or executable procedure.
}
\end{table*}

\endgroup

\subsection{Co-verified Visual Grounding}
\label{app:visual-grounding}

\paragraph{Candidate generation and semantic verification.}
SAM3 receives the open-vocabulary visual prompts and proposes candidate regions. For prompt $p_{t,j}$, candidate $k$ has mask $m_{t,j,k}$ and bounding box $b_{t,j,k}$. Equation~\ref{eq:verify} applies $f_{\mathrm v}$ to the masked crop and its prompt. The result $a_{t,j,k}$ records whether the region matches the target concept. The score $v_{t,j,k}$ records model-reported confidence. The set $\mathcal A_{t,j}$ retains all candidates and their annotations, including candidates with $a_{t,j,k}=0$.

\paragraph{Comparison with the whole image.}
After candidate verification, $\mathcal M$ compares the annotated candidates with observations of the whole image. This step checks for disagreements in object identity, attributes, counts, and relations. It is a separate comparison, not a second output of Equation~\ref{eq:verify}. Candidate records and whole-image observations jointly form $\mathcal V_t$. Local and global observations provide complementary context.

\paragraph{Using disagreement during reasoning.}
Failed candidates remain available with their failure annotations. Disagreements between local and global observations also remain available to the reasoner. The reasoner considers these records alongside relevance, confidence, and visual context. An empty candidate set indicates missing perceptual support rather than proof of object absence. Verification provides an assessment. Unresolved visual issues can trigger rechecking of the image with existing segmentation results as prompts. Unchanged prompts, candidates, and annotations are reused. Segmentation and candidate verification need not run again at every iteration.

\subsection{Memory Representations and Illustrative Entries}
\label{app:memory-examples}

The initial memory bank contains 420 entries: 296 Episodic Memory entries, 53 Semantic Memory entries, and 71 Procedural Memory entries. The serialised \textsc{type} values \texttt{Episodic}, \texttt{Semantic\_Rule}, and \texttt{Tool} implement these three conceptual categories, respectively. These counts describe the initial bank state rather than a fixed architectural capacity. The entries below are independent examples of representation and use.

\begin{table*}[t]
\caption{Representative entries from the initial memory bank. Text is shortened for readability, while titles and statistics are retained.}
\label{tab:memory-examples}
\centering
\small
\setlength{\tabcolsep}{4pt}
\renewcommand{\arraystretch}{1.12}
\begin{tabular}{@{}p{0.19\linewidth}p{0.26\linewidth}p{0.41\linewidth}@{}}
\toprule
Memory type & Stored title & Content and usage \\
\midrule
Episodic Memory &
Verify consistency between calculated values and options &
Case-linked guidance to recompute requested quantities before matching an option. Its provenance remains episodic; recorded usage outcomes include one success and one failure. \\
\midrule
Semantic Memory &
Odds ratio calculation from $2\times2$ contingency tables &
Generalized odds-ratio knowledge, $\mathrm{OR}=ad/(bc)$, where $(a,b)$ and $(c,d)$ are case/control counts in exposed and unexposed groups. Recorded usage outcomes include seven successes and two failures. \\
\midrule
Procedural Memory &
Compass traverse angle computation &
Executable operations for parsing degree-minute bearings and computing normalized interior angles. Its zero-success, five-failure history motivates reassessment of applicability and content. \\
\bottomrule
\end{tabular}
\end{table*}

Episodic Memory preserves reasoning experiences, error analyses, and corrective lessons from specific tasks. An Episodic Memory record contains \textsc{memory\_id}, \textsc{type}, \textsc{task\_query}, \textsc{original\_trajectory\_refs}, \textsc{original\_trajectory\_summary}, \textsc{memory\_items}, \textsc{stats}, and a \textsc{consolidated} flag. Each item contains a \textsc{title}, diagnostic \textsc{description}, and repair \textsc{content}; statistics include success, failure, and last-update fields. Semantic Memory is formed by clustering similar episodic memories and summarizing each cluster into reusable rules. It retains links to the source episodes and stores their corresponding reasoning steps in a dedicated field. The original episodic memories remain available after abstraction. Later reasoning can therefore use both general rules and concrete experience. Procedural Memory additionally stores a function name, parameter schema, and executable code. Usage statistics record observed outcomes associated with retrieval or execution. An episodic lesson may be phrased as general advice while remaining linked to a particular case. Conversely, semantic guidance may describe calculation steps without being an executable procedure. Provenance, abstraction, and inference-time use jointly determine these distinctions.

\paragraph{Semantic Memory reuse across questions.}
% Source: data/MMMU/memory_bank_qwen.json, meta_9d1ad1da;
% retrieval/outcome traces: traj_2efd0f33 and traj_1def06cd.
% These records establish reuse, not an episode-to-abstraction lineage.
For example, the memory bank contains the Semantic Memory entry \emph{Strictly Separate Visual Evidence from Historical Inference}. This entry distinguishes observable chart content from background knowledge. For an MMMU question about British economic interactions with India, the recorded trace retrieves this entry and notes that the GDP table does not identify commodities. It then combines the table's temporal context with historical knowledge to select \emph{Cotton}, matching the annotation. Another trace retrieves the same entry for a question about world silver flows. It combines the depicted trade routes with historical context to select the annotated answer. These cases illustrate reuse of the evidence--knowledge distinction across different scenarios and questions. They highlight the value of Semantic Memory in preserving reasoning principles that generalize beyond individual cases and guide later questions. This highlights the value of semantic memory in preserving generalizable reasoning principles and applying them across different scenarios.
\subsection{Memory Formation, Refinement, and Retrieval}
\label{app:memory-formation}

SHM stores declarative memory as $\mathcal B_{\mathrm D}^{(s)}=\mathcal B_{\mathrm E}^{(s)}\cup\mathcal B_{\mathrm S}^{(s)}$ and procedural memory as $\mathcal B_{\mathrm P}^{(s)}$. The subscripts identify episodic, semantic, and procedural entries. All three types remain retrievable. Episodic entries preserve concrete reasoning processes, error analyses, and corrective lessons. Semantic entries summarize related episodes into reusable rules. Procedural entries provide executable methods with explicit applicability conditions. Appendix~\ref{app:memory-examples} describes their stored content and examples.

\textbf{Cold-start construction.}
The system first runs EDAR on an initial set of training samples with memory retrieval disabled. It records the reasoning trajectories and obtains feedback from reference answers after each prediction. Incorrect answers and correct answers reached after multiple iterations can supply reusable experience. Valid extracted entries are organized by type. Related episodes support semantic abstraction, and recurring computational needs can support procedural memory. These steps produce the initial completed bank $\mathcal B^{(0)}$ without updating model parameters. Memory updates remain enabled during evaluation. After the current prediction is recorded, reference-answer feedback updates usage statistics and refines relevant entries. Completed revisions are available to later evaluation questions. The reference answer for the current question is not available when its prediction is generated. These results therefore evaluate reasoning with feedback-driven memory updates, while all model parameters remain fixed.

\textbf{Retrieval by title similarity.}
For question $Q_i$, the system reads $\mathcal B^{(s_i)}$. The text encoder $\phi$ maps the question to embedding $\mathbf q_i$ and the title of each memory entry $\mu_\ell$ to embedding $\mathbf z_\ell$. The encoder receives the entry's title, not its full content. Let $\sigma_i(\mu_\ell)$ denote the cosine similarity between these two embeddings. The initial candidate set is
\begin{equation}
\mathcal C_i=\operatorname{Top}_{k'_{\mathrm m}}
\!\left(\mathcal B^{(s_i)};\sigma_i\right),
\qquad k'_{\mathrm m}=\max(3k_{\mathrm m},5).
\label{eq:memory-retrieve}
\end{equation}
The operator returns up to $k'_{\mathrm m}$ entries ranked by title similarity; if the bank contains fewer entries, all available entries are returned. The candidate budget expands the pool before reranking. The constants 3 and 5 specify the existing retrieval budget; no optimality claim is attached to them. Figure~\ref{fig:sensitivity} reports sensitivity to the final retrieval count $k_{\mathrm m}$.

The ranking score $w_{i\ell}$ combines title similarity, past usage performance, and memory type. The cosine similarity is multiplied by the usage weight $\rho_\ell\in[0,1]$ and type weight $\beta_\ell$. Entries with zero usage weight are excluded before thresholding. The type weight is 1 for episodic memory, 1.15 for semantic memory, and 1.20 for procedural memory. These settings give semantic and procedural entries a $15\%$ and $20\%$ boost relative to episodic entries with the same similarity and usage weight.

The final retrieval step applies the threshold $\theta_{\mathrm m}$ to the weighted score $w_{i\ell}$ and selects up to $k_{\mathrm m}$ eligible entries. The result can be empty. The selected set $\mathcal U_i$, also denoted by $\mathcal U_Q$, stays fixed throughout the current question. The threshold $\theta_{\mathrm m}$ controls memory selection.
% AUTHOR CHECK: confirm theta_m for the reported experiments. The later code implements a 0.75 default, which does not establish the experimental setting.
Declarative memory provides knowledge and concrete experience. Retrieved procedures become available for execution when the question is applicable and all required inputs are present.

\textbf{Feedback and refinement of existing memory.}
Reference answer $y_i$ is used only after the prediction $\hat A_i$ is produced. The task protocol supplies correctness $c_i\in\{0,1\}$. A correct answer reinforces usage statistics for relevant memories. An incorrect answer prompts examination of the retrieved memories. An applicable memory with incomplete or incorrect guidance is refined by $f_{\mathrm m}$ in Equation~\ref{eq:memory-refine}. Refinement can correct the guidance, add applicability conditions, or revise a program. A valid revision replaces the corresponding entry by identifier. If no valid revision is produced, the original memory remains.

A memory can be retrieved because its title is similar to the question but still be inapplicable. Such a case calls for adjustment of its retrieval priority, not automatic rewriting of its content. Usage statistics record outcomes associated with memory use. Refining existing memories reduces redundant additions; it does not impose a fixed upper bound on memory size.

\begin{algorithm}[t]
\caption{SHM update after answering.}
\label{alg:memory-update}
\small
\begin{algorithmic}[1]
\Require Case $(Q_i,\Pi_i,\hat A_i,y_i,c_i,\tau_i)$; retrieved set $\mathcal U_i$; bank $\mathcal B^{(s)}$
\Ensure Updated bank $\mathcal B^{(s+1)}$
\State $\mathcal F_i,\mathcal F_i^{\prime},\mathcal N_i\gets\varnothing$
\State Update usage weights $\rho_\ell$ of all $\mu_\ell\in\mathcal U_i$ according to $c_i$
\If{$c_i=0$}
  \ForAll{$\mu_\ell\in\mathcal U_i$}
    \If{$\mu_\ell$ is applicable to $Q_i$}
      \State $\mu_\ell^{\prime}\gets f_{\mathrm m}(\mu_\ell,Q_i,\Pi_i,\hat A_i,y_i)$ \Comment{Eq.~\eqref{eq:memory-refine}}
      \State \textbf{if} $\mu_\ell^{\prime}$ is valid \textbf{then} add $\mu_\ell$ to $\mathcal F_i$ and $\mu_\ell^{\prime}$ to $\mathcal F_i^{\prime}$
    \Else
      \State Adjust the retrieval priority of $\mu_\ell$
    \EndIf
  \EndFor
\EndIf
\If{$\eta_i=1$} \Comment{Eq.~\eqref{eq:memory-admission}}
  \State $\mathcal N_i\gets$ valid entries extracted from the case
\EndIf
\State $\mathcal B^{(s+1)}\gets h\bigl((\mathcal B^{(s)}\setminus\mathcal F_i)\cup\mathcal F_i^{\prime}\cup\mathcal N_i\bigr)$ \Comment{Eq.~\eqref{eq:memory-operation}}
\State \Return $\mathcal B^{(s+1)}$
\end{algorithmic}
\end{algorithm}

\textbf{Selective formation of new memories.}
Equation~\ref{eq:memory-admission} requires both an empty retrieved set $\mathcal U_i$ and an informative outcome: an incorrect answer or a correct answer reached after multiple iterations. The indicator $\eta_i=1$ permits an attempt to extract new experience, not unconditional insertion. The extracted content must be valid before it is added. Reflection compares the prediction and reasoning record with the reference answer. It identifies the difficulty and a reusable correction. An incorrect prediction remains part of the failure context and is not promoted to a general fact. The title identifies the lesson, the description explains the difficulty, and the content records the guidance. When relevant memories already exist, the system prioritizes refinement over duplicate additions.

\textbf{Abstraction across related episodes.}
The system groups episodic entries by shared capability gaps and similar solution patterns. Semantic similarity in their error descriptions and repair content supports this grouping. A shared topic alone does not establish a common solution. When a coherent group contains at least $N$ unconsolidated entries, the reasoner summarizes the group into reusable semantic rules. Incidental numbers, names, and answer choices are omitted unless they define the rule's applicability. Each semantic entry retains links to its source episodes and stores their corresponding reasoning steps. The original episodes remain available after abstraction. Later reasoning can therefore use both a general rule and concrete steps from its source cases. No new semantic entry is formed when the grouping conditions are unmet.

\textbf{Procedural memory and execution.}
Recurring computational needs can motivate an executable procedure or refinement of an existing one~\citep{luo2026storage}. Construction checks the generated code and trial execution. Detected failures prompt revision. At inference time, retrieval alone does not trigger execution. The current question must fit the procedure's applicability conditions, and its required inputs must be available. Inputs can come from the question, image descriptions, or acquired visual information. Only outputs that pass execution and output checks enter $\mathcal W_t$. A reusable formula can remain semantic guidance; procedural memory stores and invokes an executable implementation.

\subsection{EDAR Reasoning and SHM Feedback Procedure}
\label{app:reasoning-procedure}

Algorithm~\ref{alg:system-details} expands the EDAR strategy in Section~\ref{sec:method-edar} and shows its post-answer handoff to SHM. Memory retrieval runs once per question. Each iteration reuses the current evidence and validated program outputs unless an action updates them. Procedural inputs can come from the question, image descriptions, or acquired visual information. Execution depends on input availability, not on a mandatory position after segmentation.

\begin{algorithm}[t]
\caption{Full EDAR procedure with SHM feedback.}
\label{alg:system-details}
\small
\begin{algorithmic}[1]
\Require $I$, $Q$; bank $\mathcal B^{(s_i)}$; threshold $\theta_{\mathrm r}$; budget $T$
\Ensure Answer $\hat A$, reasoning record $\Pi$
\State $C\gets$ description of $I$; $(\mathcal P_1,\mathcal E_1)\gets f_{\mathrm p}(Q,C)$ \Comment{Eq.~\eqref{eq:parse}}
\State $\mathcal U_Q\gets$ top-$k_{\mathrm m}$ of $\mathcal C_i$ with $w_{i\ell}\ge\theta_{\mathrm m}$ \Comment{Eq.~\eqref{eq:memory-retrieve}}
\State $\mathcal H_0\gets\varnothing$; $\mathcal W_0\gets$ validated procedure outputs
\For{$t=1$ \textbf{to} $T$}
  \If{$t=1$}
    \State $\mathcal V_1\gets\{\mathcal A_{1,j}\}_j$ and whole-image check \Comment{Eq.~\eqref{eq:verify}}
    \State $\mathcal K_1\gets g(\mathcal E_1,\mathcal O_1)$ \Comment{Eqs.~\eqref{eq:knowledge-update}, \eqref{eq:retrieval-fuse}}
  \Else
    \State Update $\mathcal V_t$ or $\mathcal K_t$ by the chosen action \Comment{reuse the rest}
  \EndIf
  \State $\mathcal W_t\gets$ still-valid outputs in $\mathcal W_{t-1}$ $\cup$ new validated outputs
  \State $\mathcal H_t\gets\mathcal H_{t-1}\cup$ reasoning on $(Q,\mathcal V_t,\mathcal K_t,\mathcal U_Q,\mathcal W_t)$
  \State $(r_t,\mathcal D_t)\gets f_{\mathrm r}(\mathcal X_t)$
  \If{$r_t\ge\theta_{\mathrm r}$ \textbf{or} $t=T$}
    \State $\tau\gets t$; \textbf{break}
  \EndIf
  \State Choose recheck or supplementation from $\mathcal D_t,\mathcal H_t$
\EndFor
\State Generate $\hat A$ from $(\mathcal V_\tau,\mathcal K_\tau,\mathcal U_Q,\mathcal W_\tau,\mathcal H_\tau)$; finalize $\Pi$
\If{memory updates are enabled \textbf{and} $y$ is available}
  \State $c\gets$ correctness of $\hat A$; run Alg.~\ref{alg:memory-update} asynchronously
\EndIf
\State \Return $(\hat A,\Pi)$
\end{algorithmic}
\end{algorithm}

\textbf{Asynchronous update and version use.}
Algorithm~\ref{alg:memory-update} summarizes post-answer updates. The current write version $s$ need not equal the read version $s_i$. Updates locate memories by their identifiers. Only current-version memories with valid replacements enter $\mathcal F_i$, matched by identifier. Their revisions form $\mathcal F_i^{\prime}$, and valid new memories form $\mathcal N_i$. Equation~\ref{eq:memory-operation} replaces the corresponding old memories $\mathcal F_i$, adds new content, and applies hierarchical organization $h$. This equation describes content replacement and formation; the usage-statistic and retrieval-priority adjustments in Algorithm~\ref{alg:memory-update} are also retained in the completed version. Completed versions become available to later questions. They do not overwrite $\mathcal U_i$ for an ongoing question. The main evaluation follows this update path after each answered question; valid content revisions are applied when the refinement or formation conditions are met.

\paragraph{Computational cost.}
Each question performs one memory retrieval. In an iteration that generates and verifies all candidates, the visual stage uses $O(mK)$ candidate verification calls. Here, $m$ is the number of active prompts and $K$ bounds the candidates per prompt. Structural and visual retrieval use $O(n+mK)$ queries, where $n$ is the number of knowledge entities. These are full-refresh bounds; later iterations may reuse earlier results and perform fewer calls. Each iteration also includes whole-image comparison as needed and one readiness assessment. Procedural execution costs depend on the retrieved program. Post-answer memory updates run asynchronously and do not block answer generation.

\subsection{Hybrid Multimodal Knowledge Retrieval}
\label{app:knowledge-retrieval}

Knowledge retrieval is part of EDAR in Section~\ref{sec:method-edar}. The structural channel queries a Neo4j-hosted MMKG using parsed entities~\citep{liu2019mmkg}. The visual channel embeds locally verified image regions with CLIP~\citep{radford2021clip} and searches the graph's image index with FAISS~\citep{johnson2019faiss}. Using the candidate records defined in Section~\ref{sec:method-grounding}, the image regions for visual retrieval are
\begin{equation}
\begin{aligned}
\mathcal O_t=\bigcup_j\bigl\{I[b_{t,j,k}]:{}&
(m_{t,j,k},b_{t,j,k},a_{t,j,k},v_{t,j,k})\\
&\in\mathcal A_{t,j},\ a_{t,j,k}=1\bigr\}.
\end{aligned}
\label{eq:retrieval-regions}
\end{equation}
The mask crop is used for candidate verification, while the bounding-box crop is used for visual retrieval. Selecting locally verified regions for retrieval does not remove failed candidates from $\mathcal V_t$.

Let $\phi_I$ be the image encoder and $\mathcal Z_I$ the indexed image embeddings. The operator $\operatorname{NN}_2$ returns up to two indexed image identifiers nearest to a query embedding. For a set of image identifiers, $\zeta$ returns the union of their linked graph entities. An empty input yields an empty entity set. The visually matched entity set is
\begin{equation}
\mathcal E_t^{\mathrm{vis}}
=\bigcup_{o\in\mathcal O_t}
\zeta\!\left(\operatorname{NN}_2\!\left(\phi_I(o);\mathcal Z_I\right)\right).
\label{eq:visual-entities}
\end{equation}
Both the parsed entities and visually matched entities support graph retrieval. Let $\Gamma$ return the selected one-hop relations and facts under the per-entity retrieval budget. The retrieval function in Equation~\ref{eq:knowledge-update} is
\begin{equation}
g(\mathcal E_t,\mathcal O_t)
=\Gamma\!\left(\mathcal E_t\cup\mathcal E_t^{\mathrm{vis}}\right).
\label{eq:retrieval-fuse}
\end{equation}
Equation~\ref{eq:knowledge-update} adds this result to $\mathcal K_{t-1}$ rather than replacing earlier knowledge. The initial retrieval uses both channels. Later queries target the missing information identified during reasoning. If no new query is issued or retrieval returns no new facts, the existing knowledge is retained. Appendix~\ref{app:hyperparams} reports the retrieval budgets.

\subsection{Readiness Assessment and Bounded Reasoning}
\label{app:readiness}

At iteration $t$, the evaluator receives $\mathcal X_t$, which contains the question, visual evidence, graph knowledge, relevant memories, available program outputs, and the reasoning record. The readiness assessment returns a score $r_t$ and unresolved issues $\mathcal D_t$. Each issue includes a category and an explanation. Breadth identifies a missing aspect of the question. Depth identifies insufficient detail. Contradiction identifies incompatible observations or claims. Relevance identifies information outside the question's scope. These descriptions guide the next action. Visual uncertainty motivates rechecking, while missing external knowledge motivates further retrieval.

Readiness estimates whether the available information supports an answer. It is not a calibrated probability that the answer is correct. The stopping iteration $\tau$ is the first iteration at which $r_t\geq\theta_{\mathrm r}$, or the maximum budget $T$ if the threshold is never reached. The system then answers using the available evidence, memories, program outputs, and reasoning history. Budget exhaustion does not establish evidence sufficiency. Missing image details and limited graph coverage can leave uncertainty unresolved.

\section{Example from the MicroQuant-VQA Benchmark}
\label{app:microquant-example}

\begin{figure}[ht]
\centering
\includegraphics[width=0.30\linewidth]{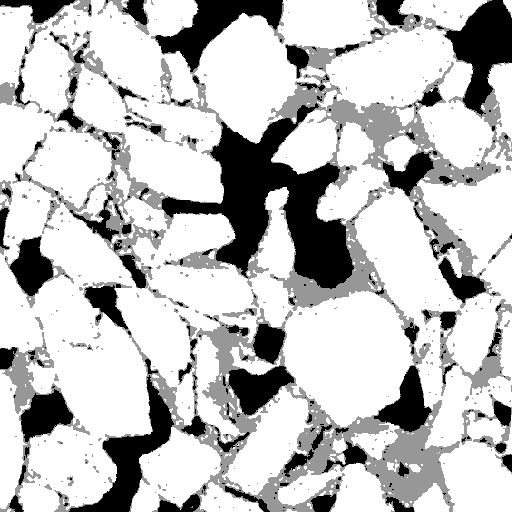}
\caption{A $512{\times}512$ three-phase segmented micro-CT slice (voxel $=$ 1.6854\,\textmu m/px) from the MicroQuant-VQA benchmark. Voxels are labeled as void (black, pixel 0), biopolymer (gray, pixel 151), or aggregate (white, pixel 255). The question asks for the lineal-path probability $L(r)$ of the aggregate phase at $r=64.01$\textmu m.}
\label{fig:microquant-example}
\end{figure}

Figure~\ref{fig:microquant-example} shows a representative instance from the MicroQuant-VQA benchmark, drawn from the lineal-path question family.

\noindent\textbf{Question.} This image is one slice of a three-phase segmented micro-CT volume (voxel $=$ 1.6854\,\textmu m/px). In this image: void $=$ black (pixel 0), biopolymer $=$ gray (pixel 151), aggregate $=$ white (pixel 255). The lineal-path function $L(r)$ of a phase is the probability that a randomly placed segment of length $r$, with both ends in the image, lies entirely within that phase. For the aggregate phase at $r = 64.01$\,\textmu m, what is the lineal-path probability?

\noindent\textbf{Options.} (A) 0.217 \quad (B) 0.2472 \quad (C) 0.0124 \quad (D) 0.147

\noindent\textbf{Ground-truth answer.} (A) 0.217 \;(full-precision value: 0.216975740131578).

\paragraph{Why this task is difficult for frozen VLMs.}
The near-chance performance of the tested Qwen3.5 backbones---$23.10\%$ for 9B and $29.30\%$ for 27B, against a $25\%$ random-guess baseline---reflects three structural challenges. First, the input is an abstract three-label segmentation mask rather than a natural image. Second, $L(r)$ requires explicit spatial measurement, which the tested models did not recover reliably by inspection. Third, every distractor is a valid same-phase value at another distance, weakening surface heuristics. Together, these properties make MicroQuant-VQA a diagnostic probe of the perception--computation gap. In our setting, grounding the target phase, computing $L(r)$ from the delineated region, and checking the requested distance were necessary for the observed improvement.

\section{Dataset Details}
\label{app:datasets}

Table~\ref{tab:dataset-statistics} summarizes the evaluated splits and task formats.

\begin{table*}[t]
\caption{Statistics and evaluation protocols for the four benchmarks.}
\label{tab:dataset-statistics}
\centering
\small
\setlength{\tabcolsep}{4pt}
\renewcommand{\arraystretch}{1.08}
\begin{tabular}{@{}lrrl@{}}
\toprule
Dataset & Available questions & Evaluated split & Answer format \\
\midrule
ScienceQA & 21{,}208 & official test & multiple choice \\
A-OKVQA & 24{,}903 & validation (1{,}145) & MC and direct answer \\
MMMU & 11{,}500 & validation (900) & multiple choice and open-ended \\
MicroQuant-VQA & 4{,}970 & validation (1{,}000) & multiple choice \\
\bottomrule
\end{tabular}
\end{table*}

\textbf{ScienceQA.}
ScienceQA~\citep{lu2022scienceqa} contains 21{,}208 questions spanning natural science, social science, and language science. Questions may include text, diagrams, or natural images. Lecture annotations are supplied as background knowledge during inference; reference rationale annotations are excluded. We evaluate the official test split and use top-1 multiple-choice accuracy. The diverse contexts make ScienceQA suitable for testing grounded perception and compositional reasoning.

\textbf{A-OKVQA.}
A-OKVQA~\citep{schwenk2022aokvqa} contains 24{,}903 questions grounded in COCO images. Its official split contains 17{,}056 training, 1{,}145 validation, and 6{,}702 test questions. Each question has four choices and ten free-form answers. We evaluate both formats on validation. Direct answers follow the official VQAv2 protocol: predictions and annotations receive the standard punctuation, digit, article, and contraction normalization, followed by leave-one-out soft accuracy with exact normalized matches.

\textbf{MMMU.}
MMMU~\citep{yue2024mmmu} contains approximately 11{,}500 college-level questions across 30 subjects and 183 subfields. It covers six disciplines and 30 heterogeneous image types, including charts, diagrams, tables, maps, and chemical structures. We evaluate all 900 validation questions. Evaluation follows the official code: option labels are parsed for multiple-choice items; open responses are reduced to candidate strings and numbers, strings use normalized containment, and numbers use equality after two-decimal normalization.

\textbf{MicroQuant-VQA.}
MicroQuant-VQA evaluates cross-domain transfer to quantitative reasoning in materials science. It is derived from three-phase segmented micro-CT slices of biopolymer-bound soil composites~\citep{miao2024bsc}. Each $512\times512$ image labels void, biopolymer, and aggregate at \(1.6854\,\mu\mathrm{m}\) per pixel. Questions require computing the lineal-path probability of a specified phase at a given distance~\citep{torquato2002random}. Each question has four options, with distractors drawn from valid same-phase values at other distances. We evaluate 1{,}000 validation questions and report top-1 multiple-choice accuracy. This benchmark complements the preceding three by focusing on numerical measurement from segmented microstructures, probing whether grounded reasoning and reusable computational procedures transfer to a specialised scientific domain.

\section{Implementation and Hyperparameters}
\label{app:hyperparams}
% AUTHOR CHECK: confirm the experimental theta_m; do not infer it from the later code default.

Table~\ref{tab:hyperparameters} reports the shared inference and memory settings. Decoding is fixed across datasets and backbone sizes. Model parameters remain fixed, while memory entries and usage statistics can change through post-answer feedback during evaluation. Retrieval budgets were selected on the ScienceQA training split and fixed before evaluation; the sensitivity sweeps in Figure~\ref{fig:sensitivity} retain the main-table configuration apart from the budget being varied. We use $\theta_{\mathrm r}=0.95$ throughout. Candidate verification returns a binary semantic match and model-reported confidence. Whole-image comparison records disagreements. Neither check introduces an additional confidence threshold in the reported procedure. The memory selection threshold $\theta_{\mathrm m}$ is distinct from the readiness threshold $\theta_{\mathrm r}$.

\begin{table*}[t]
\caption{Implementation settings used for all evaluations unless noted otherwise.}
\label{tab:hyperparameters}
\centering
\small
\setlength{\tabcolsep}{5pt}
\renewcommand{\arraystretch}{1.08}
\begin{tabular}{@{}lll@{}}
\toprule
Component & Hyperparameter & Setting \\
\midrule
Fact-reasoner decoding & temperature & $0.1$ \\
\edar & maximum iterations $T$ & $3$ \\
\edar & readiness gate $\theta_{\mathrm r}$ & $0.95$ \\
Memory retrieval & entries per query $k_{\mathrm m}$ & $2$ \\
Memory retrieval & candidate budget $k'_{\mathrm m}$ & $\max(3k_{\mathrm m},5)$ \\
Memory retrieval & episodic/semantic/procedural type weights & $1/1.15/1.20$ \\
Semantic consolidation & minimum coherent-group size $N$ & $10$ \\
MMKG retrieval & triplets per entity $k_{\mathrm g}$ & $15$ \\
FAISS retrieval & neighbors per locally supported crop & $2$ \\
\bottomrule
\end{tabular}
\end{table*}

The fact reasoner receives the image, question input, annotated evidence, type-aware memory entries, and available validated procedural outputs. When \edar is enabled, SAM3 receives prompts derived from the question and initial caption. Each candidate enters the reasoning state with its semantic judgment; failed candidates retain their failure flags. Retrieved procedures execute when their applicability conditions and input requirements are met. Inputs may come from the question, image descriptions, or acquired visual information. Only validated outputs enter the reasoning context.

\section{Details of the Comparative Methods}
\label{app:baselines}

Following the distinction in the Introduction, we group comparative methods into training-based and training-free approaches according to whether any framework component requires additional parameter training or adaptation.

\paragraph{Reporting protocol.}
Published scores follow the cited evaluation sources. ``TF'' denotes a framework requiring no additional parameter training or adaptation in any component. Off-the-shelf pretrained models are used with fixed weights throughout framework construction and inference. Our bare-backbone and incremental comparisons use identical evaluation code. QACap's occurrence-based direct-answer results remain excluded because they differ from the official VQAv2 scoring protocol used here.

\paragraph{Training-based methods.}
These methods develop reasoning capabilities through backbone fine-tuning, reinforcement learning, parameter-efficient adaptation, or auxiliary-component training. They require parameter optimization even when the answer model remains frozen.
\begin{itemize}
\item \textbf{ViLBERT} uses co-attention between visual-region and language streams. The A-OKVQA baseline fine-tunes this pretrained model on the target dataset~\citep{lu2019vilbert,schwenk2022aokvqa}.
\item \textbf{GPV-2} expands a general-purpose vision model's concept coverage with web supervision; the reported A-OKVQA configuration is subsequently fine-tuned for answering~\citep{kamath2022gpv2,schwenk2022aokvqa}.
\item \textbf{Multimodal-CoT} trains rationale generation and answer inference as two stages, integrating textual and visual features so that generated rationales condition answer prediction~\citep{zhang2024mmcot}.
\item \textbf{LLaVA and LLaVA+GPT-4 judge} use a LLaVA model fine-tuned on ScienceQA. In the ensemble, disagreements between LLaVA and text-only GPT-4 trigger a further GPT-4 call to adjudicate the final answer~\citep{liu2023llava}.
\item \textbf{CalibRL} trains Qwen2.5-VL-7B with hybrid-policy reinforcement learning using verifiable rewards, distribution-aware advantage weighting, and expert-guided calibration. Training uses geometry examples from ViRL39K; ScienceQA and MMMU are reported as out-of-domain evaluations~\citep{calibrl2026}.
\item \textbf{LLaMA-Adapter} trains visual projections and adaptation prompts with zero-initialized attention gating while freezing the image encoder and LLaMA~\citep{zhang2024llamaadapter}.
\item \textbf{LaVIN} trains mixture-of-modality adapters and routing mechanisms for visual--language instruction tuning with frozen pretrained components~\citep{luo2023lavin}.
\item \textbf{MR-MKG} trains a relational graph attention network and visual/knowledge adaptation modules while freezing the LLM and visual encoder. Training includes MMKG-grounded pretraining and task-specific adaptation. The ScienceQA result uses FLAN-UL2-19B, with 248M trainable parameters reported for the auxiliary components~\citep{lee2024mrmkg}.
\item \textbf{RG-VQA} trains a ColBERTv2 retriever to select relevant knowledge-graph triples and supplies them as generator context. The reported Bunny-8B configuration additionally uses LoRA fine-tuning; it is a retriever--generator pipeline~\citep{sravanthi2025rgvqa}.
\item \textbf{PromptCap} fine-tunes a 470M OFA captioner on GPT-3-synthesized, VQAv2-derived examples. It then prompts frozen GPT-3 with captions and in-context examples for A-OKVQA. The paper reports the code-davinci-002 model as 175B. Its validation direct-answer score is 56.3 under VQAv2 soft accuracy~\citep{hu2023promptcap}.
\item \textbf{QACap} trains a question-aware captioner with feature distillation and regional consistency learning on captions derived from KB-VQA training splits, including A-OKVQA, and passes the resulting visual descriptions to a frozen LLM~\citep{yang2025qacap}.
\item \textbf{MAGIC-VQA} combines commonsense retrieval and type-based filtering with a trained GCN/MLP that supplies answer-confidence signals to a frozen LVLM. The paper reports 0.33M trainable parameters for this auxiliary module. The A-OKVQA and MMMU scores in our tables use GPT-4o-mini; the paper does not specify its parameter count. The 0.33M figure therefore excludes the backbone~\citep{yang2025magicvqa}.
\item \textbf{V3Fusion} selects ensembles of pretrained VLMs and trains a fusion module for their outputs. Multiple-choice tasks use an MLP; open-ended tasks use a trained encoder--decoder. The A-OKVQA ensemble contains LLaVA-v1.6-Vicuna-13B, LLaVA-v1.6-Vicuna-7B, and DeepSeek-VL2-Tiny. The MMMU ensemble replaces the 13B LLaVA with Qwen2.5-VL-7B. The reported results include uncertainty-based rectification. We use the Table 2 scores of 89.17 on A-OKVQA and 56.09 on MMMU~\citep{tekin2025v3fusion}.
\end{itemize}

\paragraph{Training-free methods.}
These methods use pretrained models with fixed parameters, either through direct inference or through prompting, retrieval, and tool-based composition.
\begin{itemize}
\item \textbf{MG$^2$-RAG} builds a multimodal graph linking entities, visual regions, and document chunks, then propagates query relevance to retrieve evidence for a frozen answer model~\citep{dai2026mg2rag}.
\item \textbf{VLMs-only} use off-the-shelf GPT-4V, GPT-4o-mini, Qwen2.5-VL-7B, Gemini-3.1-Pro, and GPT-5.2 without framework-specific retrieval or adaptation. Gemini-3.1-Pro and GPT-5.2 ScienceQA scores are taken from the direct-inference evaluations reported by MG$^2$-RAG~\citep{yue2024mmmu,openai2024gpt4omini,bai2025qwen25vl,googledeepmind2026gemini,openai2026gpt52,dai2026mg2rag}.
\end{itemize}

\paragraph{Controlled internal baselines.}
The bare-VLM baseline uses the frozen fact reasoner without retrieval, SAM3, or memory. The incremental ablation first enables MMKG retrieval alone. The next stage completes EDAR by adding visual verification and adaptive reasoning to this retrieval baseline. The final stage adds SHM. Thus, the EDAR row measures the increment over its retrieval-only baseline. MMKG retrieval remains part of the complete strategy.

\section{Additional Experimental Analysis}
\label{app:additional-analysis}

\subsection{MicroQuant-VQA Transfer}
\label{app:microquant-transfer}

The main paper reports the MicroQuant-VQA comparison in Table~\ref{tab:microquant-transfer}. Dataset details and a representative question appear in Appendices~\ref{app:datasets} and~\ref{app:microquant-example}.

\subsection{Qwen3.5-27B Component Ablation}
\label{app:ablation-27b}

We repeat the incremental ablation with Qwen3.5-27B to examine whether the framework's components remain useful with a stronger fact reasoner and how their contributions change relative to the 9B setting in Table~\ref{tab:ablation}.

\begin{table}[t]
\caption{Component contributions on Qwen3.5-27B. The $+$EDAR row adds visual verification and adaptive reasoning to MMKG retrieval, as in Table~\ref{tab:ablation}. Parentheses report gains over the preceding configuration.}
\label{tab:ablation-27b}
\centering
\small
\setlength{\tabcolsep}{3pt}
\renewcommand{\arraystretch}{1.08}
\begin{tabular}{@{}p{0.40\linewidth}cc@{}}
\toprule
Configuration & ScienceQA (\%) & MMMU (\%) \\
\midrule
\raggedright Bare VLM (Qwen3.5-27B) & $91.98$ & $54.20$ \\
\raggedright \quad + MMKG retrieval & $93.00$ ($+1.02$) & $58.36$ ($+4.16$) \\
\raggedright \quad + EDAR (full strategy) & $96.11$ ($+3.11$) & $68.12$ ($+9.76$) \\
\raggedright \quad + SHM (full \method) & $\mathbf{98.56}$ ($+2.45$) & $\mathbf{75.32}$ ($+7.20$) \\
\bottomrule
\end{tabular}
\end{table}

Table~\ref{tab:ablation-27b} follows the same staged comparison. MMKG retrieval supplies external relations and facts. Completing EDAR adds visual verification and adaptive evidence acquisition. The final stage enables use and post-answer refinement of declarative and procedural memories. Each stage improves both benchmarks under this ordering. The EDAR increment is measured over retrieval alone, while the SHM increment is measured over the complete EDAR strategy.

Comparison with the 9B ablation reveals benchmark-dependent scale effects. On ScienceQA, the higher bare-27B accuracy leaves less room for improvement, consistent with the smaller absolute component gains. On MMMU, however, the gain from completing EDAR beyond retrieval alone increases from $6.85$ to $9.76$ points, and the memory gain from $5.01$ to $7.20$, while the retrieval gain remains comparable. This pattern suggests that a stronger backbone can benefit more from grounded evidence assessment and experience reuse on specialized reasoning tasks. Scaling therefore remains complementary to the framework's evidence and memory mechanisms.

\subsection{Complementarity of Declarative and Procedural Memory}
This ablation examines whether declarative and procedural memory make complementary contributions to \method. We group Episodic and Semantic Memory as declarative memory: they provide case-specific experience and generalized knowledge, respectively. Procedural Memory instead supplies executable operations for applying knowledge to the current problem. We compare no memory retrieval, declarative retrieval, and declarative plus procedural retrieval. Each configuration starts from the same initial bank and updates its own copy from post-answer feedback during evaluation. The comparison measures the effect of enabling different memory types together with their subsequent memory updates.

\begin{table*}[t]
\caption{Contributions of declarative and procedural memory on Qwen3.5-9B. Episodic and Semantic Memory jointly form the declarative category; adding Procedural Memory tests its incremental value beyond declarative retrieval. Parentheses report gains over the preceding row. All configurations start from the same initial bank and update their memories separately during evaluation.}
\label{tab:memory-ablation}
\centering
\small
\setlength{\tabcolsep}{6pt}
\renewcommand{\arraystretch}{1.05}
\begin{tabular}{@{}lcc@{}}
\toprule
Memory configuration & ScienceQA (\%) & MMMU (\%) \\
\midrule
No retrieved memory & $92.76$ & $61.29$ \\
\quad + Episodic + Semantic Memory & $94.28$ ($+1.52$) & $63.21$ ($+1.92$) \\
\quad + Episodic + Semantic + Procedural Memory & $\mathbf{95.33}$ ($+1.05$) & $\mathbf{66.30}$ ($+3.09$) \\
\bottomrule
\end{tabular}
\end{table*}

Table~\ref{tab:memory-ablation} shows higher scores with declarative retrieval and further gains with procedural retrieval. Each configuration uses its own evolving memory bank. The gains therefore reflect memory reuse and subsequent updates, rather than retrieval from identical fixed contents. Declarative memory provides prior experience and generalized knowledge, while procedural memory supplies reusable computations. The larger gain from adding procedural memory on MMMU is consistent with its specialized computational demands.

\subsection{Retrieval-Budget Sensitivity}
\begin{figure*}[t]
\centering
\includegraphics[width=0.8\linewidth]{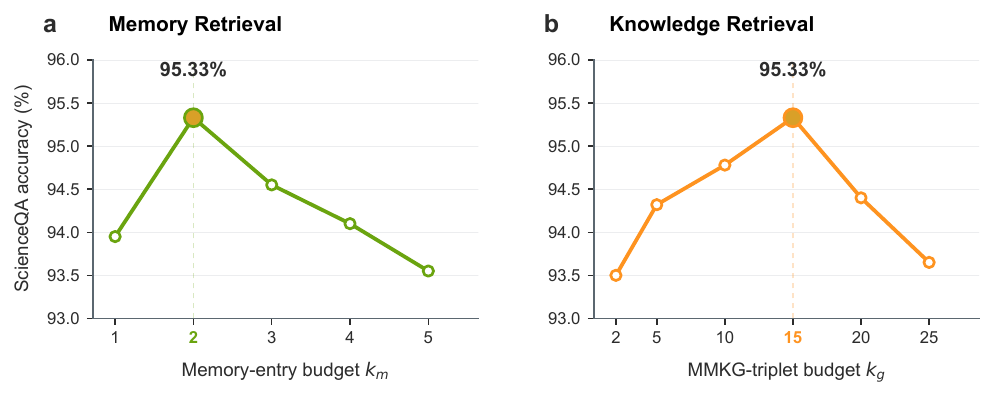}
\caption{Retrieval-budget sensitivity on ScienceQA-test with Qwen3.5-9B. Gold markers indicate settings selected on the ScienceQA training split. The highest accuracy of $95.33\%$ is achieved when $k_{\mathrm m}=2$ and $k_{\mathrm g}=15$.}
\label{fig:sensitivity}
\end{figure*}

Figure~\ref{fig:sensitivity} reveals a non-monotonic relationship between retrieval budget and answer accuracy. Within the tested ranges, performance peaks under the configuration $k_{\mathrm m}=2$ and $k_{\mathrm g}=15$, then declines when either budget is increased further. Thus, retrieving more memory entries or MMKG triplets does not necessarily improve reasoning. This pattern is consistent with a trade-off between evidence coverage and contextual selectivity: broader retrieval may introduce redundant or weakly relevant material that dilutes useful evidence. The experiment motivates bounded retrieval as a means of balancing access to external evidence with focused reasoning.

\subsection{Evidence Arbitration and Iteration Profile}
\label{sec:exp-iteration-profile}
\begin{figure}[t]
\centering
\includegraphics[width=\linewidth]{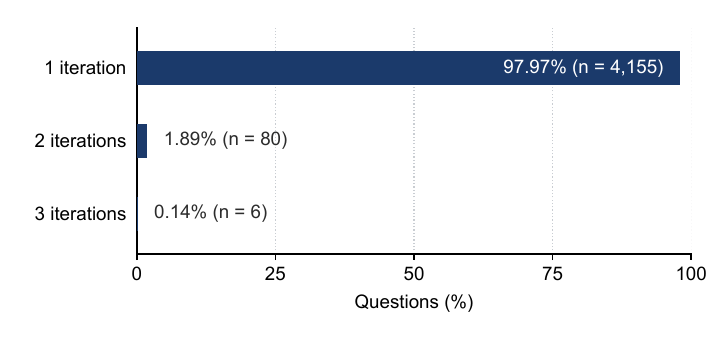}
\caption{\textbf{Selective refinement beyond the initial evidence assessment.} Under readiness-gated arbitration, most questions terminate after one outer iteration, which can already integrate perception, retrieval, and memory-informed reasoning. The small multi-iteration tail shows that further refinement is invoked selectively for unresolved cases. Results cover one Qwen3.5-9B run on the full ScienceQA test split.}
\label{fig:iteration-profile}
\end{figure}

Figure~\ref{fig:iteration-profile} examines how frequently readiness-gated arbitration invokes further refinement. Across all 4{,}241 ScienceQA test questions, $97.97\%$ terminate after one outer iteration, with a mean of $1.022$ iterations per question. The initial stage can already incorporate memory retrieval, grounded evidence acquisition, and evidence arbitration. The combination of $95.33\%$ overall accuracy and predominantly single-iteration completion highlights an iteration-efficient reasoning pattern, with readiness-guided control dynamically invoking deeper refinement only for cases with unresolved evidence needs. Table~\ref{tab:ablation} provides the complementary evidence for component-level accuracy gains; the iteration profile describes how the complete framework operates under the evaluated conditions.

\section{Discussion}
\label{sec:discussion}
\label{app:extended-discussion}

\paragraph{Why cooperative verification works.}
The incremental ablations support complementary roles for evidence arbitration and experience reuse: completing EDAR beyond retrieval alone contributes the largest added gain on both benchmarks, and SHM remains beneficial after its inclusion. \edar brings grounded candidates and conflicting observations into the fact reasoner's assessment, while SHM makes prior episodes, semantic knowledge, and procedures available across questions. This distinction connects instance-specific evidence assessment with accumulated experience. Both mechanisms can inform the initial reasoning stage, with further refinement available when the current evidence remains insufficient.

\paragraph{Why scale and the framework are complementary.}
Within Qwen3.5, the 9B-to-27B transition raised MMMU by $9.02$ points, compared with $3.23$ on ScienceQA and less than one point on each A-OKVQA format. Qwen3.6-35B further increased direct-answer and MMMU accuracy, although this comparison combines capacity and model-version changes. The stronger MMMU gain may reflect its greater dependence on latent disciplinary knowledge. Verification can protect relevant evidence from contradiction, but the fact reasoner must still possess the concepts required to interpret it.

\paragraph{Broader implications.}
Closed-loop verification may help other settings where frozen multimodal models require auditable outputs. Typed masks, boxes, and verifier scores support explicit contradiction detection. Similar mechanisms could benefit scientific documents, medical-image triage, and embodied agents, although these extensions require separate evaluation.

\clearpage
\section*{LLM Usage Statement}
% AUTHOR CHECK: name the tools previously used for English editing before submission.
The framework and experiments use pretrained Qwen3.5-9B, Qwen3.5-27B, Qwen3.6-35B, and GPT-4o-mini models as described in Section~\ref{sec:exp}, with model parameters fixed. Codex was used to edit the English grammar and phrasing of author-written text throughout the paper. Codex was used to adapt the manuscript to the official CVPR 2027 \LaTeX{} template, and Claude was used to check formatting and suggest layout edits, including the condensed Algorithm~\ref{alg:system} and short passages in Sections~\ref{sec:related} and~\ref{sec:method-edar}. The authors reviewed all edits and are responsible for all content.
{
\small
\flushbottom
\bibliographystyle{ieeenat_fullname}
\bibliography{main}
}
\end{document}

%% file: preamble.tex
\usepackage{algorithm}
\usepackage{algpseudocode}
\usepackage{multirow}
\usepackage{pifont}
\usepackage{tikz}
\newcommand{\method}{SAGE-VQA\xspace}
\newcommand{\edar}{EDAR\xspace}

%% file: authors.tex
\author{\parbox{0.95\textwidth}{\centering
Ruirui Shen\textsuperscript{1}, Jianqi Gao\textsuperscript{1}, Xiangfeng Luo\textsuperscript{1}, Yonggang Zhang\textsuperscript{2}\\
Muran Yu\textsuperscript{3}, Barney Haoyun Miao\textsuperscript{4}, Hang Yu\textsuperscript{1}, Yike Guo\textsuperscript{2}\\[6pt]
\small\textsuperscript{1} Shanghai University (shu.edu.cn)\\
\textsuperscript{2} The Hong Kong University of Science and Technology (ust.hk)\\
\textsuperscript{3} Stanford University (stanford.edu)\\
\textsuperscript{4} University of Calgary (ucalgary.ca)
}}

%% file: main.bib
@inproceedings{marino2019okvqa,
author = {Marino, Kenneth and Rastegari, Mohammad and Farhadi, Ali and Mottaghi, Roozbeh},
booktitle = {Proceedings of the IEEE/CVF Conference on Computer Vision and Pattern Recognition (CVPR)},
pages = {3195--3204},
title = {{OK-VQA}: A Visual Question Answering Benchmark Requiring External Knowledge},
year = {2019}
}

@inproceedings{schwenk2022aokvqa,
author = {Schwenk, Dustin and Khandelwal, Apoorv and Clark, Christopher and Marino, Kenneth and Mottaghi, Roozbeh},
booktitle = {European Conference on Computer Vision (ECCV)},
title = {{A-OKVQA}: A Benchmark for Visual Question Answering using World Knowledge},
year = {2022}
}

@inproceedings{lu2022scienceqa,
author = {Lu, Pan and Mishra, Swaroop and Xia, Tony and Qiu, Liang and Chang, Kai-Wei and Zhu, Song-Chun and Tafjord, Oyvind and Clark, Peter and Kalyan, Ashwin},
booktitle = {Advances in Neural Information Processing Systems (NeurIPS)},
title = {Learn to Explain: Multimodal Reasoning via Thought Chains for Science Question Answering},
year = {2022}
}

@inproceedings{yue2024mmmu,
author = {Yue, Xiang and Ni, Yuansheng and Zhang, Kai and Zheng, Tianyu and Liu, Ruoqi and Zhang, Ge and Stevens, Samuel and Jiang, Dongfu and Ren, Weiming and Sun, Yuxuan and others},
booktitle = {Proceedings of the IEEE/CVF Conference on Computer Vision and Pattern Recognition (CVPR)},
title = {{MMMU}: A Massive Multi-discipline Multimodal Understanding and Reasoning Benchmark for Expert {AGI}},
year = {2024}
}

@article{bai2025qwen25vl,
author = {Bai, Shuai and Chen, Keqin and Liu, Xuejing and Wang, Jialin and Ge, Wenbin and Song, Sibo and Dang, Kai and Wang, Peng and Wang, Shijie and Tang, Jun and others},
journal = {arXiv preprint arXiv:2502.13923},
title = {{Qwen2.5-VL} Technical Report},
year = {2025}
}

@inproceedings{gui2022kat,
author = {Gui, Liangke and Wang, Borui and Huang, Qiuyuan and Hauptmann, Alexander and Bisk, Yonatan and Gao, Jianfeng},
booktitle = {Proceedings of the 2022 Conference of the North American Chapter of the Association for Computational Linguistics (NAACL)},
pages = {956--968},
title = {{KAT}: A Knowledge Augmented Transformer for Vision-and-Language},
year = {2022}
}

@inproceedings{lin2022revive,
author = {Lin, Yuanze and Xie, Yujia and Chen, Dongdong and Xu, Yichong and Zhu, Chenguang and Yuan, Lu},
booktitle = {Advances in Neural Information Processing Systems (NeurIPS)},
title = {{REVIVE}: Regional Visual Representation Matters in Knowledge-Based Visual Question Answering},
year = {2022}
}

@inproceedings{ding2022mukea,
author = {Ding, Yang and Yu, Jing and Liu, Bang and Hu, Yue and Cui, Mingxin and Wu, Qi},
booktitle = {Proceedings of the IEEE/CVF Conference on Computer Vision and Pattern Recognition (CVPR)},
pages = {5089--5098},
title = {{MuKEA}: Multimodal Knowledge Extraction and Accumulation for Knowledge-based Visual Question Answering},
year = {2022}
}

@inproceedings{hu2023reveal,
author = {Hu, Ziniu and Iscen, Ahmet and Sun, Chen and Wang, Zirui and Chang, Kai-Wei and Sun, Yizhou and Schmid, Cordelia and Ross, David A. and Fathi, Alireza},
booktitle = {Proceedings of the IEEE/CVF Conference on Computer Vision and Pattern Recognition (CVPR)},
pages = {23369--23379},
title = {{REVEAL}: Retrieval-Augmented Visual-Language Pre-Training with Multi-Source Multimodal Knowledge Memory},
year = {2023}
}

@inproceedings{lee2024mrmkg,
author = {Lee, Junlin and Wang, Yequan and Li, Jing and Zhang, Min},
booktitle = {Proceedings of the 62nd Annual Meeting of the Association for Computational Linguistics (ACL)},
pages = {10767--10782},
title = {Multimodal Reasoning with Multimodal Knowledge Graph},
year = {2024}
}

@inproceedings{yang2025qacap,
author = {Yang, Zhen and Tao, Zhuo and Chen, Qi and Li, Liang and Qi, Yuankai and Van Den Hengel, Anton and Huang, Qingming},
booktitle = {2025 IEEE/CVF Conference on Computer Vision and Pattern Recognition (CVPR)},
doi = {10.1109/CVPR52734.2025.02305},
pages = {24753--24762},
title = {Separation of Powers: On Segregating Knowledge from Observation in {LLM}-enabled Knowledge-based Visual Question Answering},
year = {2025}
}

@article{dai2026mg2rag,
author = {Dai, Sijun and Huang, Qiang and You, Xiaoxing and Yu, Jun},
title = {{MG2-RAG}: Multi-Granularity Graph for Multimodal Retrieval-Augmented Generation},
journal = {arXiv preprint arXiv:2604.04969},
year = {2026}
}

@inproceedings{sravanthi2025rgvqa,
author = {Sravanthi, Settaluri Lakshmi and Agarwal, Pulkit and Mondal, Debjyoti and others},
booktitle = {Findings of the Association for Computational Linguistics: EMNLP},
pages = {24048--24060},
title = {{RG-VQA}: Leveraging Retriever-Generator Pipelines for Knowledge Intensive Visual Question Answering},
year = {2025}
}

@inproceedings{yang2022pica,
author = {Yang, Zhengyuan and Gan, Zhe and Wang, Jianfeng and Hu, Xiaowei and Lu, Yumao and Liu, Zicheng and Wang, Lijuan},
booktitle = {Proceedings of the AAAI Conference on Artificial Intelligence},
title = {An Empirical Study of {GPT-3} for Few-Shot Knowledge-Based {VQA}},
year = {2022}
}

@inproceedings{tiong2022pnpvqa,
author = {Tiong, Anthony Meng Huat and Li, Junnan and Li, Boyang and Savarese, Silvio and Hoi, Steven C.H.},
booktitle = {Findings of the Association for Computational Linguistics: EMNLP},
pages = {951--967},
title = {Plug-and-Play {VQA}: Zero-shot {VQA} by Conjoining Large Pretrained Models with Zero Training},
year = {2022}
}

@inproceedings{guo2023img2llm,
author = {Guo, Jiaxian and Li, Junnan and Li, Dongxu and Tiong, Anthony Meng Huat and Li, Boyang and Tao, Dacheng and Hoi, Steven C.H.},
booktitle = {Proceedings of the IEEE/CVF Conference on Computer Vision and Pattern Recognition (CVPR)},
pages = {10867--10877},
title = {From Images to Textual Prompts: Zero-shot Visual Question Answering with Frozen Large Language Models},
year = {2023}
}

@article{zhang2024mmcot,
author = {Zhang, Zhuosheng and Zhang, Aston and Li, Mu and Zhao, Hai and Karypis, George and Smola, Alex},
journal = {Transactions on Machine Learning Research (TMLR)},
title = {Multimodal Chain-of-Thought Reasoning in Language Models},
year = {2024}
}

@inproceedings{yang2025magicvqa,
author = {Yang, Shuo and Han, Caren and Luo, Siwen and Hovy, Eduard},
booktitle = {Findings of the Association for Computational Linguistics: ACL},
pages = {16967--16986},
title = {{MAGIC-VQA}: Multimodal And Grounded Inference with Commonsense Knowledge for Visual Question Answering},
year = {2025}
}

@article{tekin2025v3fusion,
author = {Tekin, Selim Furkan and others},
journal = {arXiv preprint},
title = {Vision Verification Enhanced Fusion of {VLMs} for Efficient Visual Reasoning},
year = {2026}
}

@inproceedings{hu2023avis,
author = {Hu, Ziniu and Iscen, Ahmet and Sun, Chen and Chang, Kai-Wei and Sun, Yizhou and Ross, David A. and Schmid, Cordelia and Fathi, Alireza},
booktitle = {Advances in Neural Information Processing Systems (NeurIPS)},
title = {{AVIS}: Autonomous Visual Information Seeking with Large Language Model Agent},
year = {2023}
}

@inproceedings{carion2026sam3,
author = {Carion, Nicolas and Gustafson, Laura and Hu, Yuan-Ting and Debnath, Shoubhik and Hu, Ronghang and Suris Coll-Vinent, Didac and Ryali, Chaitanya and Alwala, Kalyan Vasudev and Khedr, Haitham and Huang, Andrew and Lei, Jie and Ma, Tengyu and Guo, Baishan and Kalla, Arpit and Marks, Markus and Greer, Joseph and Wang, Meng and Sun, Peize and R\"{a}dle, Roman and Afouras, Triantafyllos and Mavroudi, Effrosyni and Xu, Katherine and Wu, Tsung-Han and Zhou, Yu and Momeni, Liliane and HAZRA, RISHI and Ding, Shuangrui and Vaze, Sagar and Porcher, Francois and Li, Feng and Li, Siyuan and Kamath, Aishwarya and Cheng, Ho Kei and Dollar, Piotr and Ravi, Nikhila and Saenko, Kate and Zhang, Pengchuan and Feichtenhofer, Christoph},
booktitle = {International Conference on Learning Representations},
editor = {C. Vondrick and B. Hariharan and C. Raffel and L. Pinto and D. Yang and A. Faust},
pages = {138846--138923},
title = {{SAM 3}: Segment Anything with Concepts},
url = {https://proceedings.iclr.cc/paper_files/paper/2026/file/e0982cbc81401df3430ee1ff780dc7a2-Paper-Conference.pdf},
volume = {2026},
year = {2026}
}

@inproceedings{liu2019mmkg,
author = {Liu, Ye and Li, Hui and Garcia-Duran, Alberto and Niepert, Mathias and Onoro-Rubio, Daniel and Rosenblum, David S.},
booktitle = {European Semantic Web Conference (ESWC)},
title = {{MMKG}: Multi-Modal Knowledge Graphs},
year = {2019}
}

@inproceedings{radford2021clip,
author = {Radford, Alec and Kim, Jong Wook and Hallacy, Chris and Ramesh, Aditya and Goh, Gabriel and Agarwal, Sandhini and Sastry, Girish and Askell, Amanda and Mishkin, Pamela and Clark, Jack and Krueger, Gretchen and Sutskever, Ilya},
booktitle = {Proceedings of the 38th International Conference on Machine Learning (ICML)},
title = {Learning Transferable Visual Models From Natural Language Supervision},
year = {2021}
}

@article{johnson2019faiss,
author = {Johnson, Jeff and Douze, Matthijs and J{\'e}gou, Herv{\'e}},
journal = {IEEE Transactions on Big Data},
number = {3},
pages = {535--547},
title = {Billion-Scale Similarity Search with {GPUs}},
volume = {7},
year = {2019}
}

@inproceedings{shinn2023reflexion,
author = {Shinn, Noah and Cassano, Federico and Berman, Edward and Gopinath, Ashwin and Narasimhan, Karthik and Yao, Shunyu},
booktitle = {Advances in Neural Information Processing Systems (NeurIPS)},
title = {Reflexion: Language Agents with Verbal Reinforcement Learning},
year = {2023}
}

@inproceedings{park2023generative,
author = {Park, Joon Sung and O'Brien, Joseph C. and Cai, Carrie J. and Morris, Meredith Ringel and Liang, Percy and Bernstein, Michael S.},
booktitle = {Proceedings of the 36th Annual ACM Symposium on User Interface Software and Technology (UIST)},
title = {Generative Agents: Interactive Simulacra of Human Behavior},
year = {2023}
}

@inproceedings{xu2025amem,
author = {Xu, Wujiang and Liang, Zujie and Mei, Kai and Gao, Hang and Tan, Juntao and Zhang, Yongfeng},
booktitle = {Advances in Neural Information Processing Systems (NeurIPS)},
title = {{A-MEM}: Agentic Memory for {LLM} Agents},
year = {2025}
}

@inproceedings{hoang2026pas,
author = {Hoang, Nhat and Vu, Minh and Thai, My T. and Bhattarai, Manish},
booktitle = {Proceedings of the IEEE/CVF Conference on Computer Vision and Pattern Recognition (CVPR)},
pages = {18273--18283},
title = {{PAS}: Prelim Attention Score for Detecting Object Hallucinations in Large Vision-Language Models},
year = {2026}
}

@inproceedings{li2026groundingme,
author = {Li, Rang and Li, Lei and Ren, Shuhuai and Tian, Hao and Gu, Shuhao and Li, Shicheng and Yue, Zihao and Wang, Yudong and Ma, Wenhan and Yang, Zhe and Ma, Jingyuan and Sui, Zhifang and Luo, Fuli},
booktitle = {Proceedings of the IEEE/CVF Conference on Computer Vision and Pattern Recognition (CVPR)},
pages = {2412--2422},
title = {{GroundingME}: Exposing the Visual Grounding Gap in {MLLMs} through Multi-Dimensional Evaluation},
year = {2026}
}

@inproceedings{liu2023llava,
author = {Liu, Haotian and Li, Chunyuan and Wu, Qingyang and Lee, Yong Jae},
booktitle = {Advances in Neural Information Processing Systems (NeurIPS)},
title = {Visual Instruction Tuning},
year = {2023}
}

@inproceedings{zhang2024llamaadapter,
author = {Zhang, Renrui and Han, Jiaming and Liu, Chris and Gao, Peng and Zhou, Aojun and Hu, Xiangfei and Yan, Shilin and Lu, Pan and Li, Hongsheng and Qiao, Yu},
booktitle = {International Conference on Learning Representations (ICLR)},
title = {{LLaMA-Adapter}: Efficient Fine-tuning of Large Language Models with Zero-initialized Attention},
year = {2024}
}

@inproceedings{luo2023lavin,
author = {Luo, Gen and Zhou, Yiyi and Ren, Tianhe and Chen, Shengxin and Sun, Xiaoshuai and Ji, Rongrong},
booktitle = {Advances in Neural Information Processing Systems (NeurIPS)},
title = {Cheap and Quick: Efficient Vision-Language Instruction Tuning for Large Language Models},
year = {2023}
}

@inproceedings{calibrl2026,
author={Zhuoxu Huang and Mengxi Jia and Hao Sun and Xuelong Li and Jungong Han},
booktitle={The Fourteenth International Conference on Learning Representations},
title={Controllable Exploration in Hybrid-Policy {RLVR} for Multi-Modal Reasoning},
year={2026},
url={https://openreview.net/forum?id=5wxyCidRsK}
}

@article{miao2024bsc,
author = {{Miao}, Barney H. and {Headrick}, Robert J. and {Li}, Zhiye and {Spanu}, Leonardo and {Loftus}, David J. and {Lepech}, Michael D.},
journal = {Cleaner Materials},
title = {Development of biopolymer composites using lignin: A sustainable technology for fostering a green transition in the construction sector},
volume = {14},
pages = {100279},
year = {2024}
}

@book{torquato2002random,
author = {Torquato, Salvatore},
title = {Random Heterogeneous Materials: Microstructure and Macroscopic Properties},
publisher = {Springer Science \& Business Media},
year = {2002}
}

@misc{googledeepmind2026gemini,
author = {{Google DeepMind Team}},
title = {Gemini 3.1 Pro: Best for Complex Tasks and Bringing Creative Concepts to Life},
year = {2026},
howpublished = {\url{https://deepmind.google/models/gemini/pro/}}
}

@misc{openai2026gpt52,
author = {{OpenAI Team}},
title = {Introducing {GPT}-5.2: The Most Advanced Frontier Model for Professional Work and Long-running Agents},
year = {2026},
howpublished = {\url{https://openai.com/index/introducing-gpt-5-2/}}
}

@inproceedings{hu2023promptcap,
author = {Hu, Yushi and Hua, Hang and Yang, Zhengyuan and Shi, Weijia and Smith, Noah A. and Luo, Jiebo},
booktitle = {2023 IEEE/CVF International Conference on Computer Vision (ICCV)},
title = {{PromptCap}: Prompt-Guided Image Captioning for {VQA} with {GPT}-3},
year = {2023},
pages = {2951--2963},
doi = {10.1109/ICCV51070.2023.00277},
publisher = {IEEE Computer Society}
}

@inproceedings{kamath2022gpv2,
author = {Kamath, Amita and Clark, Christopher and Gupta, Tanmay and Kolve, Eric and Hoiem, Derek and Kembhavi, Aniruddha},
booktitle = {Computer Vision -- ECCV 2022},
title = {Webly Supervised Concept Expansion for General Purpose Vision Models},
year = {2022},
publisher = {Springer Nature Switzerland},
pages = {662--681},
isbn = {978-3-031-20059-5}
}

@inproceedings{lu2019vilbert,
author = {Lu, Jiasen and Batra, Dhruv and Parikh, Devi and Lee, Stefan},
booktitle = {Advances in Neural Information Processing Systems (NeurIPS)},
title = {{ViLBERT}: Pretraining Task-Agnostic Visiolinguistic Representations for Vision-and-Language Tasks},
year = {2019}
}

@incollection{tulving1972episodic,
author = {Tulving, Endel},
title = {Episodic and Semantic Memory},
booktitle = {Organization of Memory},
editor = {Tulving, Endel and Donaldson, Wayne},
pages = {381--403},
publisher = {Academic Press},
address = {New York},
year = {1972}
}

@inproceedings{fang2026memp,
author = {Fang, Runnan and Liang, Yuan and Wang, Xiaobin and Wu, Jialong and Qiao, Shuofei and Xie, Pengjun and Huang, Fei and Chen, Huajun and Zhang, Ningyu},
title = {{Mem}$^p$: Exploring Agent Procedural Memory},
booktitle = {Findings of the Association for Computational Linguistics: {ACL} 2026},
pages = {17490--17502},
publisher = {Association for Computational Linguistics},
year = {2026},
doi = {10.18653/v1/2026.findings-acl.866},
url = {https://aclanthology.org/2026.findings-acl.866/}
}

@inproceedings{luo2026storage,
author = {Luo, Jinghao and Tian, Yuchen and Cao, Chuxue and Luo, Ziyang and Lin, Hongzhan and Li, Kaixin and Kong, Chuyi and Yang, Ruichao and Ma, Jing},
title = {From Storage to Experience: A Survey on the Evolution of {LLM} Agent Memory Mechanisms},
booktitle = {Findings of the Association for Computational Linguistics: {ACL} 2026},
pages = {41622--41652},
publisher = {Association for Computational Linguistics},
address = {San Diego, California, United States},
year = {2026},
doi = {10.18653/v1/2026.findings-acl.2069},
url = {https://aclanthology.org/2026.findings-acl.2069/}
}

@article{wang2023voyager,
author = {Wang, Guanzhi and Xie, Yuqi and Jiang, Yunfan and Mandlekar, Ajay and Xiao, Chaowei and Zhu, Yuke and Fan, Linxi and Anandkumar, Anima},
title = {Voyager: An Open-Ended Embodied Agent with Large Language Models},
journal = {arXiv preprint arXiv:2305.16291},
year = {2023}
}

@misc{openai2024gpt4omini,
  author = {{OpenAI}},
  title = {{GPT-4o mini}: Advancing Cost-efficient Intelligence},
  year = {2024},
  howpublished = {\url{https://openai.com/index/gpt-4o-mini-advancing-cost-efficient-intelligence/}}
}

@inproceedings{ouyang2026reasoningbank,
author = {Ouyang, Siru and Yan, Jun and Hsu, I-Hung and Chen, Yanfei and Jiang, Ke and Wang, Zifeng and Han, Rujun and Le, Long T. and Daruki, Samira and Tang, Xiangru and Tirumalashetty, Vishy and Lee, George and Rofouei, Mahsan and Lin, Hangfei and Han, Jiawei and Lee, Chen-Yu and Pfister, Tomas},
title = {{ReasoningBank}: Scaling Agent Self-Evolving with Reasoning Memory},
booktitle = {International Conference on Learning Representations (ICLR)},
year = {2026},
url = {https://openreview.net/pdf/24a0425aa87cfcd0ba7344bbf5e11cdfe9e497f4.pdf}
}

@article{han2023survey,
author = {Han, Kai and Wang, Yunhe and Chen, Hanting and Chen, Xinghao and Guo, Jianyuan and Liu, Zhenhua and Tang, Yehui and Xiao, An and Xu, Chunjing and Xu, Yixing and Yang, Zhaohui and Zhang, Yiman and Tao, Dacheng},
title = {A Survey on Vision Transformer},
journal = {IEEE Transactions on Pattern Analysis and Machine Intelligence},
volume = {45},
number = {1},
pages = {87--110},
year = {2023},
doi = {10.1109/TPAMI.2022.3152247}
}

@article{brauwers2023general,
author = {Brauwers, Gianni and Frasincar, Flavius},
title = {A General Survey on Attention Mechanisms in Deep Learning},
journal = {IEEE Transactions on Knowledge and Data Engineering},
volume = {35},
number = {4},
pages = {3279--3298},
year = {2023},
doi = {10.1109/TKDE.2021.3126456}
}

@article{sumers2024coala,
author = {Sumers, Theodore R. and Yao, Shunyu and Narasimhan, Karthik and Griffiths, Thomas L.},
title = {Cognitive Architectures for Language Agents},
journal = {Transactions on Machine Learning Research},
year = {2024},
url = {https://openreview.net/forum?id=1i6ZCvflQJ}
}
